\documentclass[10pt,twocolumn,letterpaper]{article}

\usepackage{cvpr}      

\definecolor{cvprblue}{rgb}{0.21,0.49,0.74}
\usepackage[pagebackref,breaklinks,colorlinks,allcolors=cvprblue]{hyperref}
\usepackage{url}
\usepackage{lipsum}
\usepackage{graphicx} 
\usepackage{mwe}
\usepackage{subcaption}
\usepackage{booktabs}
\usepackage{multirow}
\usepackage[capitalize]{cleveref}
\usepackage{colortbl}
\usepackage[inline]{enumitem}

\newcommand{\rri}{\includegraphics[width=0.025\linewidth]{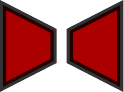}}
\newcommand{\gri}{\includegraphics[width=0.025\linewidth]{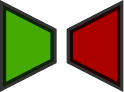}}
\newcommand{\ggi}{{\includegraphics[width=0.025\linewidth]{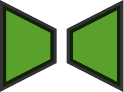}}}
\newcommand{\rgi}{{\includegraphics[width=0.025\linewidth]{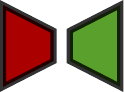}}}
\newcommand{\ok}{{\includegraphics[width=0.04\linewidth]{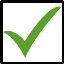}}}
\newcommand{\notok}{{\includegraphics[width=0.04\linewidth]{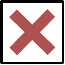}}}

\usepackage{colortbl}

\def\paperID{12953} 
\def\confName{CVPR}
\def\confYear{2025}

\title{Revisiting MAE pre-training for 3D medical image segmentation}
\author{Tassilo Wald \thanks{Equal contribution. Authors are permitted to list their name first in their CVs.}$^{~~1,2,3}$,\quad Constantin Ulrich$^{*~1,4}$, \\ Stanislav Lukyanenko$^{6}$,\quad Andrei Goncharov$^{6}$, Alberto Paderno$^{6,7}$, \quad Maximilian Miller$^{6}$, \\ Leander Maerkisch$^{6}$, \quad Paul Jaeger$^{8,9}$, \quad Klaus Maier-Hein$^{1,2,3,4,5,9}$\\
\small $^{1}$~Division of Medical Image Computing, German Cancer Research Center (DKFZ); $^{2}$~Helmholtz Imaging, DKFZ\\
  \small$^{3}$~Faculty of Mathematics and Computer Science,
  University of Heidelberg; $^{4}$~Medical Faculty Heidelberg, University of Heidelberg\\
 \small$^{5}$~ National Center for Tumor Diseases (NCT), NCT Heidelberg; $^{6}$~ FLOY, Munich, Germany \\ \small $^{7}$~Department of Biomedical Sciences, Humanitas University (Milan), IRCCS Humanitas Research Hospital, Rozzano, Italy.\\
    \small$^{8}$~Interactive Machine Learning Group, DKFZ Heidelberg; $^{9}$~Pattern Analysis and Learning Group, Department of Radiation Oncology \\ 
{\tt\small tassilo.wald@dkfz-heidelberg.de}
}

\begin{document}
\maketitle
\begin{abstract}
Self-Supervised Learning (SSL) presents an exciting opportunity to unlock the potential of vast, untapped clinical datasets, for various downstream applications that suffer from the scarcity of labeled data.
While SSL has revolutionized fields like natural language processing and computer vision, its adoption in 3D medical image computing has been limited by three key pitfalls: Small pre-training dataset sizes, architectures inadequate for 3D medical image analysis, and insufficient evaluation practices. In this paper, we address these issues by i) leveraging a large-scale dataset of 39k 3D brain MRI volumes and ii) using a Residual Encoder U-Net architecture within the state-of-the-art nnU-Net framework. iii) A robust development framework, incorporating 5 development and 8 testing brain MRI segmentation datasets, allowed performance-driven design decisions to optimize the simple concept of Masked Auto Encoders (MAEs) for 3D CNNs. The resulting model not only surpasses previous SSL methods but also outperforms the strong nnU-Net baseline by an average of approximately 3 Dice points setting a new state-of-the-art.
Our code and models are made available \href{https://github.com/MIC-DKFZ/nnssl}{here}.
\end{abstract}    
\section{Introduction}
\label{sec:introduction}

\begin{figure}
    \centering
\includegraphics[width=.75\linewidth]{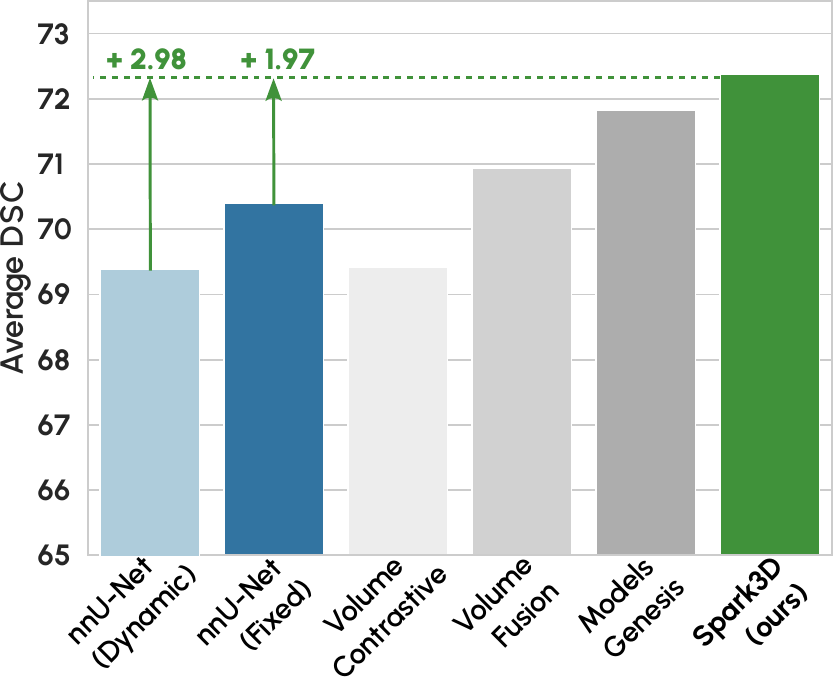}
    \caption{\textbf{Well-configured MAE pre-training for CNNs is state-of-the-art.} When comparing our MAE pre-trained model on all eleven test Datasets, Spark3D achieves almost +3 DSC points compared to the strong nnU-Net baseline and outperforms current SSL methods.}
    \label{fig:figure_1_average_performance}
\end{figure}

In recent years, the concept of Self-Supervised  Learning (SSL) has emerged as a driving factor in data-rich domains, enabling large-scale pre-training that facilitates the learning of robust and transferable general-purpose representations \citep{assran2023self,oquab2023dinov2,he2022masked}.
This paradigm shift has been instrumental in advancing various fields, particularly in domains with abundant labeled data like NLP or natural vision. In the domain of 3D medical image computing, this trend has not caught on. 
\noindent Currently, the domain is either focused on training from-scratch, mainly using the nnU-Net framework by \citet{isensee2021nnu}, or using supervised pre-training, which is limited by the cost associated with annotated data \citep{wasserthal2023totalsegmentator,ulrich2023multitalent,huang2023stu}. The usage of supervised pre-training implies the willingness of the domain to adopt pre-training and calls to question the currently established SSL methods in the domain.  
We believe this lack of widespread adoption of previously established methods can be attributed to three major pitfalls of previous SSL research in this field:

\textbf{P1 - Limited dataset size:} Many SSL approaches have been developed on very few unlabeled volumes, often being trained on fewer than 10,000 images \cite{zhou2021models, wu2024voco, tang2024hyspark, tang2022self, zhuang2023advancing, hatamizadeh2021swin,gu2024self, chaitanya2020contrastive}, almost approaching scales of supervised dataset sizes.
These datasets tend to be pooled from publicly available annotated datasets, as larger datasets pose a greater hurdle to acquire. While many hospitals possess 3D medical images in the millions, they are locked away from the public due to patient privacy concerns.
Some larger open-source datasets exist, e.g.,  the Adolescent Brain Cognitive Development (ABCD) dataset of the NIH (N=40k)~\citep{national2015adolescent} or the UK Biobank dataset (N=120k), but they restrict access pending an internal review board's approval, posing a hurdle for open science. 

\textbf{P2 - Outdated Backbones:} Many studies develop their SSL method on non-state-of-the-art architectures, e.g., utilizing transformers \citep{wu2024voco, tang2022self, wang2023mis, chen2023masked, gu2024self, chaitanya2020contrastive}.
While transformers are prevalent in the 2D natural imaging domain \citep{dosovitskiy2020image}, recent architectures leveraging attention \citep{vaswani2017attention} have so far not been able to reach state-of-the-art performance in 3D medical segmentation. 
In fact, well-configured 3D U-Net \citep{cciccek20163d, ronneberger2015u} inspired CNNs dominate 3D medical image segmentation, outperforming transformer-based models by a large margin \cite{isensee2024nnu}.
This underscores the need for SSL methods that can be seamlessly integrated with CNNs to harness their full potential in medical image analysis on downstream tasks.


\textbf{P3 - Insufficient Evaluation:} Existing methods often lack rigorous evaluation, masking the methods' efficacy.
This is represented through:
i) Evaluating too few datasets to show generalization of the pre-training \citep{valanarasu2023disruptive, chaitanya2020contrastive}
ii) Stacking multiple contributions, e.g., novel architecture with new pre-training, which does not allow one to draw conclusions if the pre-training is effective without the architecture \citep{wang2023mis}
iii) Comparing to bad from-scratch baselines, like a badly configured, outdated \citet{cciccek20163d} UNet instead of a well-configured 3D nnU-Net CNN baseline \citep{isensee2021nnu}. 
iv) Evaluating their method on seen data they pre-trained on.
We want to emphasize that we do not intend to point fingers, but to raise awareness that evaluation matters and insufficient evaluation can lead to a lack of clarity which methods are the best. 
This observation is similar to the recent study by \citet{isensee2024nnu}, which shows that this is also prevalent in the field of medical image segmentation when training from scratch.

\begin{table}[t]
    \centering
    \caption{\textbf{Current SSL methods are limited by three pitfalls:} They are trained on less than 10K 3D images (P1), they do not use a state-of-the-art backbone for 3D image segmentation as recommended by \citet{isensee2024nnu} (P2) or they do not evaluate their method on more than 4 downstream datasets (P3). \ok~indicates evading a pitfall while \notok~indicates falling for a pitfall.}
    \label{tab:pitfalls}
    \resizebox{\linewidth}{!}{
    \begin{tabular}{lr|ccc}
    \toprule
    \multirow{2}{*}{Method} & \multirow{2}{*}{Year} & P1 & P2 & P3\\ 
     &    &  Data & Model & Eval.\\
     \hline
    3DMAE \citep{chen2023masked}   &  Jan. 2023 & \notok & \notok  & \notok \\
    GL-MAE \citep{zhuang2023advancing}   &  Aug. 2023 & \notok & \notok  & \ok \\
    MAPSeg \citep{zhang2024mapsegunifiedunsuperviseddomain}   & March 2023 & \notok & \notok  & \notok \\
    HySparK \citep{tang2024hyspark}   &  Aug. 2024 & \notok &  \ok  & \ok \\
    AMAES \citep{munk2024amaes} & Aug. 2024 & \ok & \ok & \notok \\
    Swin UNETR \citep{tang2022self} & March 2022 & \notok & \notok &  \ok\\
    VoCo \citep{wu2024voco}& April 2024 & \notok & \notok & \ok \\
    VF \citep{wang2023mis} & June 2023 & \ok & \notok & \notok \\
    MG \citep{zhou2021models} & Aug. 2019 & \notok & \notok & \ok \\
    GVSL \citep{he2023geometric} & March 2023 & \notok & \notok & \ok \\
    \midrule
    S3D (ours) &  & \ok & \ok & \ok  \\
    \bottomrule
    \end{tabular}}
    
\end{table}

\noindent In this paper, we explore the Masked Auto Encoder (MAE) paradigm for 3D CNNs, with the recent adaptations introduced by \citet{tian2023designing, woo2023convnext}. Our approach addresses the key pitfalls of previous SSL research in 3D medical image segmentation, ensuring broader applicability and higher performance. Our main contributions are as follows:
\begin{enumerate}
     \item \textbf{State-of-the-Art Performance:} Our final model achieves superior performance compared to existing SSL-pre-trained models using the same architecture and pre-training dataset, see \cref{fig:figure_1_average_performance}. Notably, this is the first work to demonstrate that SSL pre-training with a fixed architecture can consistently outperform a state-of-the-art, dynamically optimized nnU-Net baseline by approximately 3 Dice points across 11 diverse downstream datasets \cite{isensee2021nnu}.
    \item \textbf{Comprehensive Evaluation Addressing Previous Pitfalls:} Building on the detailed analysis of limitations in prior work (see \cref{tab:pitfalls}), we set up a robust development environment to design our method. 
    \begin{enumerate*}[(\roman*)]
        \item P1: We utilize a collection of 39k 3D MRI volumes for self-supervised pre-training, substantially larger than most datasets used in prior SSL work \cite{zhou2021models, wu2024voco, tang2024hyspark, tang2022self, zhuang2023advancing, hatamizadeh2021swin,valanarasu2023disruptive}.
        \item P2: By using the state-of-the-art Residual Encoder U-Net CNN architecture \cite{isensee2024nnu}, we bypass the limitations of outdated or transformer-based architectures, leveraging CNNs’ superior performance in 3D medical segmentation.
        \item P3: We rigorously evaluate our model across five development and eight testing datasets, covering a wide range of downstream tasks, diverse pathologies, novel image modalities, and multi-center datasets. Moreover, we disentangle the pre-training methodology from other contributions like novel backbones and include the strong 3D nnU-Net baseline, ensuring reliable results.

    \end{enumerate*}
        \item \textbf{In-Depth Ablation Studies:} We provide detailed analyses of critical design decisions, such as masking ratio, sparsification, and fine-tuning strategies. Additional ablations explore generalization, low-data performance, and faster fine-tuning, offering valuable insights.  \\
\end{enumerate}
\noindent While the fundamental concept of an MAE is not novel, its application in 3D medical image analysis has yet to be fully validated due to limitations in previous studies, which we illustrate in \cref{tab:pitfalls}. Our carefully designed experiments close these gaps and reveal MAEs true potential.

\section{Development Framework}
\label{sec:development_framework}






The goal of this paper is to develop a robust SSL pre-training method. 
Due to the limited prior work in the 3D medical domain, many design choices need to be made. We address this by sequentially validating each methodological contribution on five downstream development datasets before testing the final configuration on eight untouched test datasets. 
To reduce the search space and disentangle the effects of SSL pre-training from other basic design choices, we choose to keep some parameters fixed based on best practices in the domain:\\
\begin{enumerate*}[label=(\textbf{\roman*})]
\item The architecture used is always the same state-of-the-art residual encoder U-Net architecture (ResEnc U-Net) \citep{isensee2024nnu,ulrich2023multitalent}.
\item The input patch size is set to [160x160x160].
\item All images are resampled to the same target spacing of [1x1x1] mm \citep{roy}.
\item All images are z-score normalized to zero mean and unit variance \citep{isensee2021nnu}.
\item As optimizer, we use SGD with a decreasing poly-learning rate \citep{poly} following nnU-Net \citep{isensee2021nnu}.
\item We always employ random sampling during pre-training, irrespective of the prevalence of the different MR modalities.
\end{enumerate*}

\paragraph{Pre-training Dataset}
To develop our pre-training method, we utilize a proprietary brain MRI dataset sourced from over 44 centers containing over 9k patients comprising a total of approx. 44k 3D MRI scans.
Due to the variety of data sources, this dataset contains images from more than 10 different MR scanners, various MR modalities and a diverse patient population. For more details on the distributions, we refer to \Cref{fig:datasetfigure}.

\noindent Since this data is sourced directly from clinical examinations, it includes empty or broken images, poor-quality images and so-called scout scans used to determine the field of view of the patient in the MR. 
Since these scans are not used in diagnostics, these images are filtered by discarding
\begin{enumerate*}[label=(\alph*)]
    \item images with a field of view of $<50mm$ in any axis,
    \item images with a spacing $>6.5mm$ in any direction and
    \item images of file size $<200kb$, which indicate an empty image.
\end{enumerate*}
Moreover due to the low quantity of MR Angiography, Susceptibility weighted Images (SWI) and Proton Density (PD) weighted images, we restrict our training data to only include T1, T2, T1 FLAIR and T2 FLAIR images, resulting in our final pre-training dataset of 39,168 MR images. 

\paragraph{Development Datasets}
After pre-training we fine-tune on five datasets and calculate the average Dice Similarity Coefficient (DSC) per dataset to evaluate the effectiveness of the pre-training. The multiple datasets are essential to ensure that our design choices do not overfit to a specific MRI modality or pathological target.
Specifically, we utilize: 
\begin{enumerate}
    \item \textit{MS FLAIR (D1)}: Consensus delineations of multiple sclerosis (MS) lesions on T2-weighted FLAIR images~\citep{MUSLIM2022108139}.
    \item \textit{Brain Mets (D2)}: Brain metastases imaged through T1, contrast enhanced gradient echo T1ce, contrast enhanced spin echo T1 and a T2 FLAIR sequence acquired in the Stanford University Hospital \citep{grovik2020deep}.
    \item \textit{Hippocampus (D3)}: The Hippocampus dataset, task 4 of the medical segmentation decathlon (MSD) \citep{antonelli2022medical}, contains delineations of the anterior and posterior Hippocampus in T1 weighted MRI \citep{simpson2019large}. 
    \item \textit{Atlas22 (D4)}: Anatomical Tracings of Lesions After Stroke (ATLAS) on T1 weighted images. We use the Atlas R2.0 dataset from \citet{liew2022large}.
    \item \textit{CrossModa (D5)}: Delineations of intra-meatal and extra-meatal vestibular schwannoma tumors and cochlea delineations in contrast-enhanced T1 weighted MRI \citep{dorent2023crossmoda}. 
\end{enumerate}
Of all these datasets, we set aside a hold-out test set comprising 20\% of all images before the start of the method development process.
The remaining images were split 80/20 into training and validation sets for the development process.

\paragraph{Test Datasets}
Additionally, eight hold-out test datasets were used to evaluate the efficacy of our learned representations when fine-tuning them to segment other target structures.
\begin{enumerate}
    \item \textit{Cosmos (D6):} This dataset features carotid vessel wall segmentation and atherosclerosis diagnosis, of which we use the contours to evaluate segmentation performance \citep{cosmos22challenge}.
    \item \textit{HaNSeg (D7):} This dataset contains segmentations of 30 organs at risk (OAR), with associated T1 MRI and CT, of which we only use the  MR images for model development \citep{podobnik2023han}.
    \item \textit{Isles22 (D8):} This dataset contains annotations of ischemic stroke lesions, with associated diffusion-weighted imaging (DWI), apparent diffusion coefficient (ADC) and T2 FLAIR \citep{hernandez2022isles}.  
    \item \textit{HNTS-MRG24 (D9):} T2-weighted MR images of pre-treatment oropharyngeal cancer and metastatic lymph nodes with associated annotations \citep{hntsmrg2024wahid}.
    \item \textit{BraTS Africa (D10):} This dataset contains 1.5 Tesla T1, T1ce, T2, and T2 FLAIR MRIs of glioblastoma and high-grade gliomas imaged in Nigeria. 
    \item \textit{T2 Aneurysms (D11):} This proprietary dataset contains 240 T2 MRI images of segmented brain aneurysms with the surrounding brain tissue.
    \item \textit{TOF Angiography Aneurysms (D12):} This proprietary dataset contains 144 time-of-flight MR-angiography images of segmented brain aneurysms with the surrounding brain tissue.
    \item \textit{BraTS Mets (D13):} This dataset holds brain metastases segmentations on T1ce MRIs, similar to \textit{D2}. However, instead of fine-tuning on it we use it to measure generalization when inferring models trained on \textit{D2} on it. 
\end{enumerate}
For test datasets \textit{D6-D12} we use an 80/20 split for fine-tuning and for testing. A validation set is omitted as we fine-tune each method only once without any interventions. For \textit{D13} all data is used for testing, as no training is conducted. \textit{D12} and \textit{D13} are only used to assess generalization performance, with \textit{D13} only appearing in Appendix \cref{tab:generalization}.

\section{Related Work}
In this section, we revisit currently established MAE pre-training paradigms. Then, we survey existing MAE pre-training paradigms in the 3D medical domain, before expanding this to a broader set of pre-training paradigms for 3D medical image segmentation. 

\subsection{Masked Autoencoders}
Auto-encoders have been established for a long time through the early pioneering works of \citet{vincent2010stacked} but recently garnered significant attention through \citet{he2022masked}, showing that a simple MAE pre-training paradigm is scalable and provides substantial performance improvements.
Subsequently, follow-up works extended these findings to video data \citep{feichtenhofer2022masked, tong2022videomae} as well as language-image data \citep{li2023scaling}. These earlier successes primarily relied on Transformer architectures, as their sequence modeling offered significant VRAM savings. 
CNNs on the other hand are less suited to such a pre-training paradigm as masking disrupts the 2D data-structure they compute on. Despite this \citet{woo2023convnext} and \citet{tian2023designing} were able to extend MAE pre-training successfully to CNNs. \citet{woo2023convnext} introduced their fully convolutional masked autoencoder (FCMAE) framework, while \citet{tian2023designing} introduced adaptations to the convolution computation and normalization to make it resilient to masking.

\subsection{Masked Autoencoder In 3D Medical}
The earliest work using autoencoders for self-supervised pre-training in the medical domain was Models Genesis (MG) conducted by \citet{zhou2021models} in 2019. They introduced reconstruction of in-painting, out-painting, and denoising of local intensity shifts as their pre-training task. Many approaches followed, with Swin UNETR by \citet{tang2022self} proposing a joint in-painting, rotation, and contrastive objective for Swin Transformer pre-training. \citet{chen2023masked} proposed a basic masking and reconstruction strategy applied to 3D Vision Transformers (ViT) \cite{dosovitskiy2020image} termed (3DMAE) and \citet{zhuang2023advancing} proposed to use a combination of masked global and local views in conjunction with a reconstruction and consistency loss to pre-train 3D ViTs, named GL-MAE. MAPSeg by \citet{zhang2024mapsegunifiedunsuperviseddomain} combines a masked autoencoder objective and pseudo-labeling to address domain shifts in a federated setting. 
More recently, in concurrent work, \citet{munk2024amaes} proposed pre-training CNNs with a vanilla masking and reconstruction objective without any adaptations on a large brain dataset (AMAES) and \citet{tang2024hyspark} presented a sparse MAE with reconstruction objective pre-trained on abdomen CT data (HySparK). 

\subsection{Other Pre-Training For 3D Medical}

Additionally, self-supervised learning approaches for 3D medical images include \citet{zhuang2019self}, who introduced a training objective aimed at recovering the orientation and rotation of shuffled and rotated sub-volumes,  inspired by a Rubik's cube. \citet{wang2023mis} proposed a method that involves mixing two images at varying patch rates and predicting the mixing categories via a segmentation objective, termed Volume Fusion (VF). Geometric Visual Similarity Learning (GVSL) of \citet{he2023geometric} leveraged anatomical consistency by learning to register two images of the same field of view and using local similarity as a training objective. More recently, \citet{wu2024voco} suggested partitioning a volume into non-overlapping base-crops and generating additional overlapping sub-crops, with a training objective to predict the degree of overlap between sub-crops and base-crops (VoCo).

\section{Revisiting 3D MAEs}
\label{sec:mae_design}

\begin{table}[t]
\centering
\caption{\textbf{Development Experiments:} During development, we evaluate the Average DSC on all development datasets to quantify the best configuration. Methods of the same configuration are denoted through common colors.}
    \begin{subtable}[t]{\linewidth}
        \centering 
        \caption{\textbf{Sparsification:} The introduction of all sparsification components showed the best results, with most improvements coming from adding the additional densification conv layer between encoder and decoder.}
        \resizebox{\linewidth}{!}{
\begin{tabular}{l|lllll|l}
\toprule
Configuration &     D1 &    D2 &   D3 &      D4 &     D5 &         Avg. D1-D5 \\
\midrule
Base               &              49.96 &              72.62 &     \textbf{89.03} &     \textbf{63.45} &  \underline{81.67} &              71.35 \\
+ Sparse Conv. + BN. &  \underline{50.34} &  \underline{73.34} &              88.91 &  \underline{62.96} &              81.25 &              71.36 \\
+ w/ Mask Token      &              50.02 &              73.21 &  \underline{88.92} &              62.84 &     \textbf{81.83} &  \underline{71.37} \\
\rowcolor{lightgray!60}+ w/ Dens. Conv.     &     \textbf{51.02} &     \textbf{74.07} &              88.91 &              62.81 &              81.50 &     \textbf{71.66} \\
\bottomrule
\end{tabular}}

        \label{subtab:sparse_ablation}
    \end{subtable}\\
    \hfill
        \begin{subtable}[t]{\linewidth}
            \centering
            \caption{\textbf{Masking ratio:} Masking ratios between 60\% and 75\% worked best when choosing static ratios. However, a dynamic range, including higher masking ratios, performs similarly and was chosen.}
            \resizebox{\linewidth}{!}{
\begin{tabular}{l|lllll|l}
\toprule
Mask ratio &     D1 &    D2 &   D3 &       D4 &     D5 &    Avg. D1-D5 \\

\midrule
30 \%         &              50.25 &              71.37 &              88.90 &              63.18 &  \underline{81.70} &              71.08 \\
45  \%        &              50.60 &              70.83 &  \underline{88.97} &  \underline{63.27} &     \textbf{81.73} &              71.08 \\
60  \%        &              50.62 &              73.56 &              88.96 &     \textbf{63.37} &              81.48 &              71.60 \\
\rowcolor{lightgray!60} 75    \%      &  \underline{51.02} &     \textbf{74.07} &              88.91 &             62.81 &              81.50 &     \textbf{71.66} \\
90   \%       &              50.56 &              72.51 &     \textbf{88.98} &              62.41 &              81.49 &              71.19 \\
\rowcolor{blue!10} U[60\%-90\%]       &     \textbf{51.49} &  \underline{74.01} &              88.83 &              62.39 &              81.54 &  \underline{71.65} \\
\bottomrule
\end{tabular}}

            \label{subtab:masking_ratio} 
        \end{subtable}\\
        \begin{subtable}[t]{\linewidth}
            \centering
            \caption{\textbf{Development performance:} When comparing our final S3D model against the baseline methods, trained equally, S3D exceeds all baselines on the development datasets.
            }
            \label{subtab:dev_performance} 
            \resizebox{\linewidth}{!}{\begin{tabular}{l|lllll|l}
\toprule
Pre-training &         D1            &      D2              &          D3          &        D4            &             D5       &    Avg. D1-D5                \\
\midrule
 No Dyn. &              45.56 &              \underline{72.26} &              \underline{88.80} &              60.44 &     \textbf{82.61} &              69.93 \\
 No Fixed &              49.37 &              69.13 &              88.78 &              60.74 &              81.33 &              69.87 \\
  VoCo &              50.35 &              67.20 &              88.22 &              57.82 &              80.29 &              68.77 \\
       VF &              49.93 &              69.58 &              \textbf{88.83} &              61.75 &              81.48 &              70.31 \\
       MG &  \underline{50.50} &              71.14 &              \textbf{88.83} &              \textbf{63.29} &              \underline{82.15} &              \underline{71.18} \\\midrule
     \rowcolor{blue!10} \textbf{S3D (ours)} &     \textbf{51.49} &     \textbf{74.01} &     \textbf{88.83} &  \underline{62.39} &  81.54 &     \textbf{71.65} \\
\bottomrule
\end{tabular}}

        \end{subtable}
    
    \label{tab:development_results}
\end{table}
Masked autoencoders (MAEs) are a well-established pre-training paradigm in the natural imaging domain and in the medical image segmentation domain for transformers. 
In this section, we investigate this paradigm and optimize it for 3D medical image segmentation using a ResEnc U-Net architecture of \citet{isensee2024nnu}.


\paragraph{Default parameters}
MAEs are trained by masking an input image to a certain degree and training the network to reconstruct the occluded regions, minimizing deviations between reconstruction and the original image.
In our experiments, we train the MAE with an L2-Loss in the z-score normalized voxel space and only calculate the reconstruction loss where regions were masked. Moreover, we do not remove skip-connections, following the general consensus of \citet{woo2023convnext}, \citet{tian2023designing}, and \citet{he2022masked}. The default hyperparameters (as used by the model denoted in gray in \cref{tab:development_results}) are learning rate 1e-2, weight decay 3e-5, batch size 6, SGD optimizer with Nesterov momentum 0.99, masking ratio of 75\% trained with a PolyLR schedule for 250k steps (this represent 1000 epochs in the nnU-Net framework) and minor spatial augmentations of affine scaling, rotation and mirroring.
\paragraph{Sparsification}
When masking the input image, CNNs are not able to ignore the masked regions in the same manner as transformers can. To address this, \citet{tian2023designing} proposed to adapt the CNN architectures to better fit the sparse inputs:
\begin{enumerate*}[label={(\textbf{\alph*})}]
    \item \textbf{Sparse Convolutions and Normalization:} Through the receptive field of convolutions masked-out regions are iteratively eroded from their boundaries. By re-applying the masked regions after every convolution this problem can be resolved. Moreover, the masks can introduce a problematic shift in the normalization layer statistics due to the introduced zero values. To resolve this normalization is constrained to only consider the non-masked values. 
    \item \textbf{Mask Token:} Instead of feeding the light-weight decoder feature maps with zeroed mask regions, the regions are densified by filling them with a learnable mask token, simplifying the reconstruction task of the decoder. 
    \item \textbf{Densification Convolution:} After filling the masks with the Mask Token and before passing the feature maps to the decoder, a [3x3x3] convolution is applied to the feature maps at every resolution except the highest resolution to prepare the representations for decoding.
\end{enumerate*}

\noindent Results of these changes are visualized in \cref{subtab:sparse_ablation}. The adaptations are introduced iteratively, meaning the 'MaskToken' ablation is only applied together with the Sparse Convolutions and Normalization. 
It can be observed that the full set of adaptations improves  performance by an average of $0.3$ DSC points across our development datasets. Subsequently, all adaptations are kept and the following evaluations are presented with these changes applied.

\paragraph{Masking strategy}
The masked region is determined by sampling randomly in the CNN's bottleneck of shape [5x5x5] and up-sampling these regions to the input resolution, to ensure the masks align in the bottleneck of the CNN architectures. This results in masking regions of [32x32x32] voxels of non-overlapping regions in the input. 
As a sampling strategy, we follow random masking, as previous work showed no benefit of structured masking for images nor videos \citep{he2022masked, feichtenhofer2022masked}.
In the scope of the development phase, we explore 5 static masking ratios between 30\% and 90\% and evaluate a dynamic masking ratio randomly masking between 60\% and 90\%.

\noindent Results are presented in \cref{subtab:masking_ratio} and highlight that the masking ratios of 60\%, 75\%, and the dynamic masking ratio of 60\% to 90\% perform equally well.
Due to the highly similar performance, we choose to proceed with a dynamic masking ratio over the static masking ratio, due to expecting this masking to be more difficult to learn. We refer to this model as Spark3D (S3D) going forward.

\begin{table*}[t]
    \centering
     \caption{\textbf{Fine-tuning matters.} We compare various combinations of weight transfer and fine-tuning schedules. 
    Transfer: \ggi~Transfer all weights, \gri~Transfer encoder weights only. Warm-Up and Fine-tuning: \rgi~Only decoder weights adapted during fine-tuning, \ggi~Encoder and decoder weights adapted during fine-tuning. 
    $\ddagger$: nnU-Net default (Dynamic planning.)}
    \label{tab:finetune_ablation}
    \resizebox{0.875\linewidth}{!}{
    
\begin{tabular}{lllll|rrrrr|l}
\toprule
Transfer & 1. Warm-Up & 2. Warm-Up & Fine-tuning & Max. LR &     D1 &  D2 &  D3 &  D4 &  D5 &   Avg  \\
\midrule
\rri   & - & - & \ggi & 1e-2 &          45.56 &            72.26 &             88.80 &         60.44 &           82.61 & 69.93$^\ddagger$ \\
 \rri & - & - & \ggi &1e-2 &          49.37 &            69.13 &             88.78 &         60.74 &           81.33 & 69.87 \\\midrule

 \ggi & - & - & \ggi &1e-2 &          50.37 &            70.64 &             88.61 &         61.51 &           81.91 & 70.61 \\
 \ggi & -  &  -   & \ggi &1e-3 &          49.98 &            71.04 &             88.68 &         61.45 &           82.12 & 70.65 \\\midrule
 \ggi & -  & \ggi & \ggi &1e-2 &          49.84 &            72.56 &             88.45 &         62.16 &           81.75 & 70.95 \\
 \ggi & -  & \ggi   & \ggi &1e-3 &          51.54 &            72.74 &             88.85 &         62.44 &           82.33 & \underline{71.58} \\
 \ggi & -  & \ggi    & \ggi &1e-4 &          50.66 &            72.98 &             88.68 &         62.73 &           82.09 & 71.43 \\\midrule 
 \gri & - & - & \ggi &1e-2 &          50.81 &            68.55 &             88.52 &         60.73 &           80.77 & 69.87 \\
 \gri & - & -     & \ggi &1e-3 &          50.11 &            73.90 &             88.58 &         61.82 &           81.48 & 71.18 \\\midrule
 
 \gri & - & \rgi & \ggi & 1e-2 &          51.39 &            72.06 &             88.65 &         61.70 &           81.78 & 71.12 \\
 \gri & - &   \rgi  & \ggi & 1e-3 &          51.04 &            72.72 &             88.91 &         62.84 &           81.19 & 71.34 \\\midrule
  \gri & \rgi & \ggi & \ggi &1e-2 &          48.81 &            72.92 &             88.84 &         62.46 &           82.04 & 71.02 \\
 \gri &  \rgi &   \ggi  & \ggi &1e-3 &          51.42 &            72.84 &             89.09 &         63.30 &           82.15 & \textbf{71.76} \\
 \gri & \rgi  &  \ggi   & \ggi &1e-4 &          50.13 &            72.26 &             88.72 &         61.70 &           81.93 & 70.95 \\
\bottomrule
\end{tabular}
}
\end{table*}

\paragraph{Fine-tuning strategy}
Given a pre-trained model, a crucial question arises: Which weights to transfer and how to schedule the fine-tuning? We investigated various different schedules, with all of them implemented in the nnU-Net framework \cite{isensee2021nnu}.
Regarding weight transfer we investigated transferring \begin{enumerate*}[label=(\roman*)]
    \item both the encoder and decoder \ggi, or
    \item only the encoder \gri ~with a randomly initialized decoder.
\end{enumerate*} 
Regarding the fine-tuning schedule, we investigate whether to use a learning rate warm-up of 12.5k steps, ramping the learning rate up to the maximum LR. When only transferring the encoder, an additional warm-up of only the decoder is investigated to adapt the randomly initialized decoder to the pre-trained encoder. In some configurations, this results in two learning rate warm-ups of 12.5k steps each.
Additionally, we investigate whether to keep the encoder frozen \rgi~for the warm-up process or to fine-tune the encoder weights as well. Lastly, we investigate whether to decrease the peak learning rate to 1e-3, 1e-4 or to keep it at the default of 1e-2. 

\noindent Results are presented in \Cref{tab:finetune_ablation} and allow three important observations to be made: \begin{enumerate*}[label=(\roman*)]
\item \textbf{Warm-up stages are essential:} Not applying a warm-up step significantly reduces performance. Including a warm-up for both the encoder and decoder boosts accuracy by 0.6 to 1 DSC points.
\item \textbf{Learning rate adjustments matter:} Reducing the peak learning rate to 1e-3 during fine-tuning consistently yields better results than the default 1e-2, with the best performance seen when fine-tuning both the encoder and decoder with lower learning rates.
\item \textbf{Freezing encoder weights is detrimental:} The encoder should not remain fixed during fine-tuning. Allowing the encoder to be fine-tuned improves the model’s performance compared to when only the decoder is fine-tuned, as observed in early exploratory work.
\end{enumerate*}



\section{Results and Discussion}
\label{sec:results}
\begin{figure}[t]
    \centering
    \includegraphics[width=\linewidth]{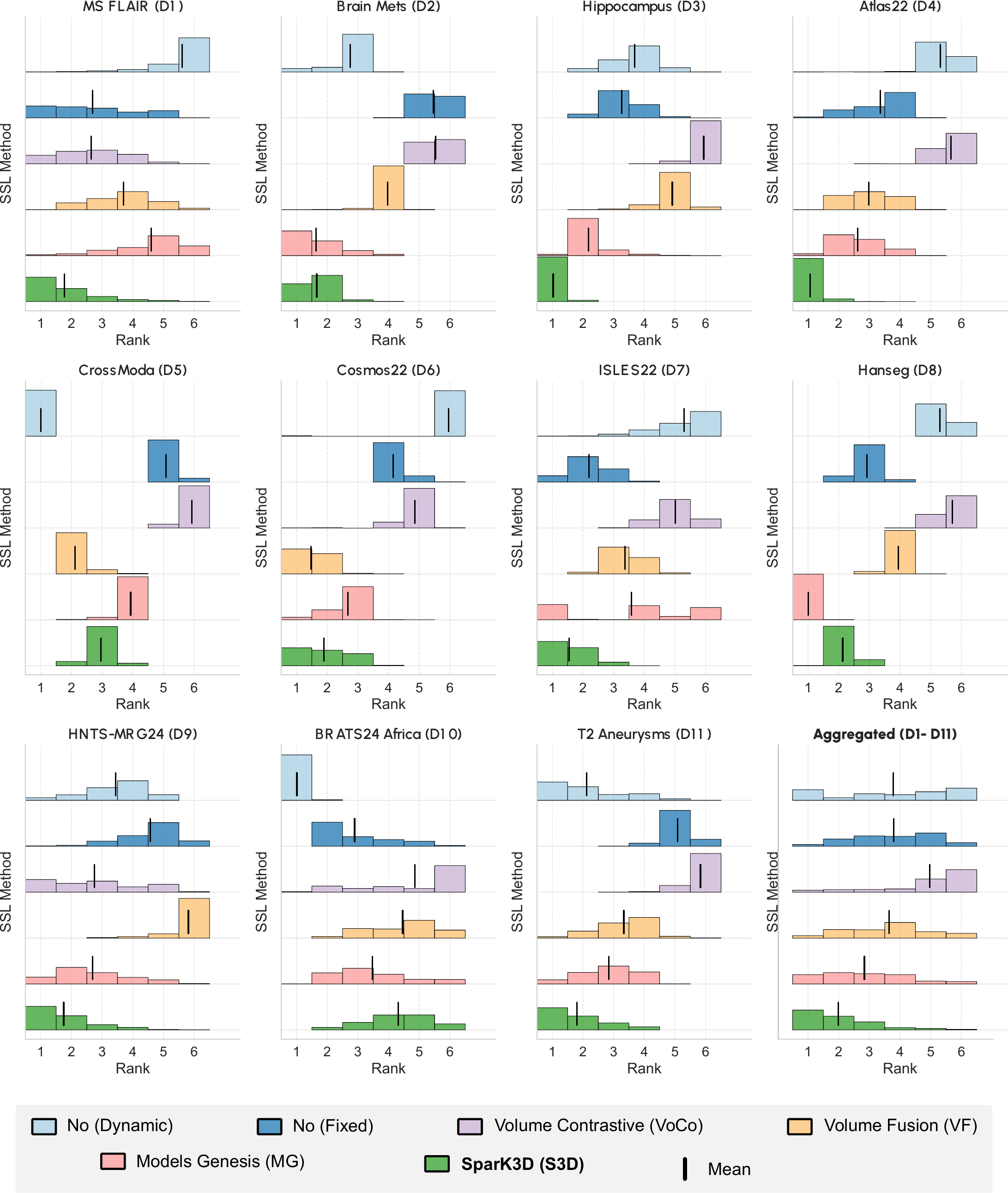}
    \caption{\textbf{S3D ranks best across all methods.} In addition to absolute mean performance, we report the ranking stability of the methods through bootstrapping for all test datasets as well as the aggregated rank across all datasets.} 
    \label{fig:test_rank_bootstrapped}
\end{figure}

We compare our final model S3D against Volume Contrastive (VoCo)~\citep{wu2024voco}, VolumeFusion (VF)~\citep{wang2023mis}, as well as  Models Genesis (MG)~\citep{zhou2021models}.
The baselines are pre-trained for 250k steps using the same framework on the same data with the same backbone and the same hyperparameters (where possible) but scaled to fully utilize an A100 40GB GPU. We provide explicit baseline methods and configuration details in \cref{apx:detailed_method_config}.
Moreover, we compare against two strong from-scratch baselines. The first, \textit{No (Dyn.)}, represents a from scratch default nnU-Net 3D architecture that was planned and preprocessed on each downstream dataset individually, potentially resulting in different architectures, data preprocessing, and spacings. The second from-scratch baseline, \textit{No (Fixed)}, is a nnU-Net training with the same plans and pre-processing as defined by our pre-training. The dataset-wise mean dice similarity coefficient (DSC) and mean normalized surface distance (NSD) at 2 mm tolerance values across our test dataset suite are provided in \cref{tab:test_performance_all}. Additionally, the ranking stability is evaluated through bootstrapping and is provided in \cref{fig:test_rank_bootstrapped}.

\subsection{Observations}
\paragraph{SSL pre-training works} Across all tested datasets, SSL pre-trained methods demonstrate improved downstream segmentation performance. Comparing our S3D method to the most similar from-scratch baseline \textit{No (Fixed)}, we observe higher DSC scores in 10 out of 11 test datasets, with an average increase of +2 DSC points and +1.6 NSD points. This improvement is not limited to our method; MG and VF also achieve higher performance than the baseline, indicating the utility of SSL methods when applied to sufficient data and a state-of-the-art architecture.\\

\noindent \textbf{MAEs dominate }
Throughout our test dataset pool, SSL schemes using the masked image modeling paradigm (MG and S3D) consistently rank higher than the contrastive VoCo or the pseudo-segmentation-based VolumeFusion pre-training method for CNN pre-training. Given the age of Models Genesis - published in 2019 - it is surprising to see it outperform the more recent VoCo or VF. We attribute this to a combination of two factors: \begin{enumerate*}
    \item Models Genesis was originally published and trained on an outdated 3D-UNet \citep{cciccek20163d} and outside of the powerful nnU-Net framework \citep{isensee2021nnu}. This highlights the importance of avoiding Pitfall 2: Training on a state-of-the-art backbone.
    \item VoCo and VF were introduced in conjunction with architectures they were optimized for. By transferring them to a CNN setting, hyperparameters chosen to optimize the method for their original architecture pre-training combination may be suboptimal for the new CNN backbone.
\end{enumerate*}

\begin{table}[t]
    \centering
        \caption{\textbf{S3D out-performs all baselines:} The upper table presents the DSC results, while the lower table displays the NSD results across the test datasets, covering a diverse range of brain MRI tasks. \textit{No Fixed} represents a from-scratch baseline sharing the same architecture, preprocessing and downstream training steps as all SSL methods. \textit{No Dyn.} represents the original nnU-Net adapted to each downstream dataset individually.}
    \label{tab:test_performance_all}
\resizebox{\linewidth}{!}{
\begin{tabular}{l|ccccc|c}
\toprule
SSL Method &            No (Dyn.) &           No (Fix.) &   VoCo &                 VF &                 MG &                \textbf{S3D} \\
\midrule
\textbf{Dataset}              &    \multicolumn{5}{c}{\textbf{Dice Similarity Coefficient (DSC)}}\\
\midrule
\rowcolor{lightgray!30}MS FLAIR (D1)        &              57.81 &  \underline{59.82} &  59.70 &              59.29 &              58.64 &     \textbf{60.35} \\
Brain Mets (D2)      &              63.66 &              56.53 &  56.25 &              61.01 &     \textbf{65.39} &  \underline{65.24} \\
\rowcolor{lightgray!30}Hippocampus (D3)     &              89.18 &              89.24 &  88.78 &              89.03 &  \underline{89.38} &     \textbf{89.60} \\
Atlas22 (D4)         &              63.28 &              65.52 &  62.97 &              65.76 &  \underline{65.93} &     \textbf{66.95} \\
\rowcolor{lightgray!30}CrossModa (D5)       &     \textbf{85.64} &              83.44 &  83.07 &  \underline{84.24} &              83.91 &              84.08 \\
Cosmos22 (D6)        &              60.28 &              78.17 &  77.40 &     \textbf{80.09} &              79.67 &  \underline{80.00} \\
\rowcolor{lightgray!30}ISLES22 (D7)         &              77.94 &  \underline{79.44} &  78.14 &              78.96 &              78.85 &     \textbf{79.70} \\
Hanseg (D8)          &              59.00 &              61.85 &  57.47 &              61.49 &     \textbf{62.52} &  \underline{62.11} \\
\rowcolor{lightgray!30}HNTS-MRG24 (D9)      &              66.73 &              65.90 &  67.65 &              63.34 &  \underline{68.00} &     \textbf{68.62} \\
BRATS24 Africa (D10) &     \textbf{93.07} &  \underline{92.51} &  91.97 &              92.16 &              92.36 &              92.19 \\
\rowcolor{lightgray!30}T2 Aneurysms (D11)   &  \underline{46.76} &              41.97 &  40.16 &              44.96 &              45.48 &     \textbf{47.26} \\ \midrule
Avg. DSC             &              69.40 &              70.40 &  69.41 &              70.94 &  \underline{71.83} &     \textbf{72.37} \\
Avg. Rank            &               4.64 &               4.55 &   6.27 &               4.36 &   \underline{3.18} &      \textbf{2.00} \\\midrule

\textbf{Dataset}              &      \multicolumn{5}{c}{\textbf{Normalized Surface Distance (NSD)}}\\ 
\midrule
\rowcolor{lightgray!30}MS FLAIR (D1)        &           78.77 &     \textbf{80.16} &              79.70 &  79.57 &              79.16 &  \underline{80.03} \\
Brain Mets (D2)      &           80.72 &              76.72 &              72.77 &  79.20 &  \underline{81.51} &     \textbf{82.53} \\
\rowcolor{lightgray!30}Hippocampus (D3)     &  \textbf{99.46} &              99.42 &              99.43 &  99.46 &              99.39 &  \underline{99.46} \\
Atlas22 (D4)         &           70.52 &              73.77 &              70.15 &  73.67 &  \underline{74.22} &     \textbf{75.35} \\
\rowcolor{lightgray!30}CrossModa (D5)       &  \textbf{99.85} &              99.76 &              99.72 &  99.78 &              99.74 &  \underline{99.81} \\
Cosmos22 (D6)        &           72.60 &              96.47 &              94.48 &  96.89 &  \underline{96.95} &     \textbf{97.45} \\
\rowcolor{lightgray!30}ISLES22 (D7)         &           88.55 &  \underline{90.45} &              89.39 &  90.28 &              89.59 &     \textbf{90.59} \\
Hanseg (D8)          &           82.20 &     \textbf{85.94} &              80.44 &  85.29 &  \underline{85.94} &              85.80 \\
\rowcolor{lightgray!30}HNTS-MRG24 (D9)      &           71.83 &              71.26 &  \underline{73.47} &  67.88 &              73.22 &     \textbf{74.07} \\
BRATS24 Africa (D10) &  \textbf{95.66} &  \underline{95.36} &              94.95 &  94.94 &              95.33 &              95.06 \\
\rowcolor{lightgray!30}T2 Aneurysms (D11)   &  \textbf{62.24} &              55.56 &              51.79 &  58.97 &              59.38 &  \underline{61.18} \\
\midrule
Avg. NSD             &           82.04 &              84.08 &              82.39 &  84.17 &  \underline{84.95} &     \textbf{85.58} \\
Avg. NSD Rank        &            4.27 &               4.27 &               5.82 &   4.64 &   \underline{4.00} &      \textbf{2.18} \\
\bottomrule
\end{tabular}

}

\end{table}

\paragraph{Impact of dynamic configuration}
Comparing the \textit{No (Dyn.)} and \textit{No (Fixed)} configurations, both trained from scratch, reveals that selecting the appropriate configuration for each dataset can significantly influence performance.
For instance, on datasets D2 and D11, the dynamic configuration outperforms the fixed by +7 and +5 DSC points, respectively, while for D6, the fixed configuration yields results +18 DSC points higher. In the majority of datasets where the fixed configuration underperforms relative to the dynamic nnU-Net, pre-training helps to recover performance. However, in some cases, such as with D5, the dynamic default nnU-Net still proves superior.

\subsection{Ablation experiments}
\paragraph{Low-Data Regime}
While previous experiments focused on comparing SSL pre-training against a from-scratch baseline with full-scale datasets available, many applications in the medical domain have access to only a very small amount of labeled images. To measure the benefits of pre-training in such a low-data regime, we artificially reduced the total amount of data available for training to 10, 20, 30, or 40 labeled images.

\noindent Results are presented in \cref{tab:low_data}.
Our pre-trained S3D model leads to better downstream performance compared to training from scratch in this setting. With just 40 trained images, the fine-tuned model nearly matches the performance of the from-scratch model trained on the full dataset.
Future research could explore whether optimizing the training duration or learning rate schedule could prevent the pre-trained network from overfitting during fine-tuning on such a limited number of training images.

\paragraph{Generalization Performance}
To assess the generalization capability of the proposed pre-training method, we tested two scenarios. 
First, we evaluated fine-tuning our method on an unseen modality using the TOF Angiography Aneurysms dataset (D12). As shown in \cref{tab:generalization}
without pre-training, the fixed configuration suffers a performance drop of 20 Dice points. We attribute this to the significant difference in median spacing for the downstream task ($[0.50, 0.43, 0.43]$ mm), which has a higher resolution than the fixed target spacing of $[1, 1, 1]$ mm used in the pre-training experiments.
This lower resolution likely increases the difficulty of segmenting small aneurysms.
Despite this decrease, pre-training mitigates some of this degradation and proves highly beneficial compared to training the same configuration from scratch.
Interestingly, Models Genesis achieves the best results, potentially due to its use of intensity augmentations during pre-training, which increases robustness against brightness shifts, such as when generalizing to different MRI sequences.
Second, we fine-tuned on the D2 dataset using only the T1 contrast-enhanced (T1ce) sequence and applied these models directly to D13 without any additional fine-tuning. While the dynamic configuration performed best on the in-distribution validation cases of D2, the results on D13 indicate that pre-training improves generalization across different centers, with our S3D model yielding the best performance.

\paragraph{Pre-training Time}
Previous studies have demonstrated the positive impact of extended training schedules on the quality of learned representations for downstream tasks \citep{he2022masked, feichtenhofer2022masked}. To explore the relevance of this factor in the 3D medical domain, we conducted a similar experiment, evaluating training durations ranging from 62.5k to 1M steps. Our results indicate that the benefits of longer training schedules begin to degrade after 250k steps, as evident in \cref{tab:pretrainind_length}. 

\paragraph{Fine-tuning length}
Initializing from pre-trained weights has the potential to reduce the computational resources needed for the network to adapt to new tasks. To assess this, we tested different fine-tuning durations on our development datasets. While maintaining 12.5k iterations for both the decoder and full network warm-up phases, we experimented with varying subsequent training lengths. As shown in \cref{tab:finetune_results_table}, adding just 12.5k additional iterations (37.5k total) already outperforms training from scratch. However, achieving optimal performance still requires completing the full fine-tuning schedule.

\paragraph{Multi-Channel Input}
In many medical examinations, it is common to perform multiple scans, as clinicians often require images with different characteristics for accurate decision-making. Consequently, some datasets, like D2, D8, and D10, contain multiple input modalities. While pre-training may involve all modalities, we only feed one modality at a time into the network since not all patients have scans in every modality. This raises the question of how to handle datasets with multiple registered images.
To address this, we conducted a 5-fold cross-validation on the D2 development dataset. We evaluated the replication of each input modality along with random initialization of the input stem weights. Additionally, we tested freezing the stem weights during the decoder warm-up phase. As shown in \cref{tab:stemtable}, the most stable and consistently effective approach was replicating the pre-trained stem and keeping it frozen during the decoder's warm-up period.

\section{Conclusion}
This work is the first to demonstrate the potential of properly configured MAEs in 3D medical image segmentation. By overcoming key pitfalls in previous research, such as limited dataset sizes, outdated architectures, and insufficient evaluation, we show a consistent performance improvement over previous SSL methods. Notably, for the first time, we achieve consistent improvements over the dynamic, dataset-adaptive nnU-Net baseline, validated across a large and diverse set of development and testing datasets. 

\paragraph{Limitations} In our current manuscript, we justify our selection of baseline methods by excluding prior work that was specific to architectures (3DMAE), methods that had no public code (GL-MAE), and methods that do not outperform a nnU-Net baseline trained from scratch (see Appendix \cref{tab:btcvresults}).
While our findings are promising, several avenues remain open for future exploration. 
We limited our method to Head and Neck MRI images, allowing us to focus on this region with a comparatively large pretraining dataset. Expanding to encompass the entire body across multiple imaging modalities would have required substantially more data to achieve comparable performance.
Notably, in our ablations, we tested increasing training time and batch size (Appendix \cref{tab:placeholderscaling}) which did not lead to performance improvements, but the question remains of whether scaling the pre-training dataset size or the model parameters could enable further improvements. Furthermore, the intensity shifts employed by Model Genesis SSL task hint at intriguing possibilities for improving generalization across unseen MRI modalities, which needs to be further explored for MAEs.
Lastly, a data-centric approach to curating the most relevant data for SSL represents an exciting frontier for future research. Raw clinical datasets often contain images not intended for diagnostic purposes, such as those used for scanner calibration, which can dilute the effectiveness of pre-training. While we applied basic filtering to exclude low-quality data, more sophisticated filtering techniques could significantly enhance the quality of the pre-training process.

\noindent To conclude, this work follows the spirit of prior studies like nnU-Net \cite{isensee2021nnu} showing that a robust development strategy, informed model configuration, and rigorous validation leads to true and sustainable performance improvements, contrasting the current hype for employing and modifying the latest network architecture. With this paper, we hope to contribute to a cultural shift in the community toward validation-driven development enabling true scientific progress.

\section{Acknowledgments}
This work was partly funded by the Helmholtz Foundation Model Initiative (HFMI) under the subproject \textit{`The Human Radiome Project'} (THRP). Part of this work was funded by Helmholtz Imaging (HI), a platform of the Helmholtz Incubator on Information and Data Science.

\newpage
{
    \small
    \bibliographystyle{ieeenat_fullname}
    \bibliography{main}

\begin{thebibliography}{50}
\providecommand{\natexlab}[1]{#1}
\providecommand{\url}[1]{\texttt{#1}}
\expandafter\ifx\csname urlstyle\endcsname\relax
  \providecommand{\doi}[1]{doi: #1}\else
  \providecommand{\doi}{doi: \begingroup \urlstyle{rm}\Url}\fi

\bibitem[Antonelli et~al.(2022)Antonelli, Reinke, Bakas, Farahani, Kopp-Schneider, Landman, Litjens, Menze, Ronneberger, Summers, et~al.]{antonelli2022medical}
Michela Antonelli, Annika Reinke, Spyridon Bakas, Keyvan Farahani, Annette Kopp-Schneider, Bennett~A Landman, Geert Litjens, Bjoern Menze, Olaf Ronneberger, Ronald~M Summers, et~al.
\newblock The medical segmentation decathlon.
\newblock \emph{Nature communications}, 13\penalty0 (1):\penalty0 4128, 2022.

\bibitem[Assran et~al.(2023)Assran, Duval, Misra, Bojanowski, Vincent, Rabbat, LeCun, and Ballas]{assran2023self}
Mahmoud Assran, Quentin Duval, Ishan Misra, Piotr Bojanowski, Pascal Vincent, Michael Rabbat, Yann LeCun, and Nicolas Ballas.
\newblock Self-supervised learning from images with a joint-embedding predictive architecture.
\newblock In \emph{Proceedings of the IEEE/CVF Conference on Computer Vision and Pattern Recognition}, pages 15619--15629, 2023.

\bibitem[Chaitanya et~al.(2020)Chaitanya, Erdil, Karani, and Konukoglu]{chaitanya2020contrastive}
Krishna Chaitanya, Ertunc Erdil, Neerav Karani, and Ender Konukoglu.
\newblock Contrastive learning of global and local features for medical image segmentation with limited annotations.
\newblock \emph{Advances in neural information processing systems}, 33:\penalty0 12546--12558, 2020.

\bibitem[Chen et~al.(2022)Chen, Zhao, Dou, Du, Yang, Sun, Yu, Zhao, Yuan, and Balu]{cosmos22challenge}
H. Chen, X. Zhao, J. Dou, C. Du, R. Yang, H. Sun, S. Yu, H. Zhao, C. Yuan, and N. Balu.
\newblock Carotid vessel wall segmentation and atherosclerosis diagnosis challenge, 2022.

\bibitem[Chen et~al.(2018)Chen, Papandreou, Kokkinos, Murphy, and Yuille]{poly}
Liang-Chieh Chen, George Papandreou, Iasonas Kokkinos, Kevin Murphy, and Alan~L. Yuille.
\newblock Deeplab: Semantic image segmentation with deep convolutional nets, atrous convolution, and fully connected crfs.
\newblock \emph{IEEE Transactions on Pattern Analysis and Machine Intelligence}, 2018.

\bibitem[Chen et~al.(2023)Chen, Agarwal, Aggarwal, Safta, Balan, and Brown]{chen2023masked}
Zekai Chen, Devansh Agarwal, Kshitij Aggarwal, Wiem Safta, Mariann~Micsinai Balan, and Kevin Brown.
\newblock Masked image modeling advances 3d medical image analysis.
\newblock In \emph{Proceedings of the IEEE/CVF Winter Conference on Applications of Computer Vision}, pages 1970--1980, 2023.

\bibitem[{\c{C}}i{\c{c}}ek et~al.(2016){\c{C}}i{\c{c}}ek, Abdulkadir, Lienkamp, Brox, and Ronneberger]{cciccek20163d}
{\"O}zg{\"u}n {\c{C}}i{\c{c}}ek, Ahmed Abdulkadir, Soeren~S Lienkamp, Thomas Brox, and Olaf Ronneberger.
\newblock 3d u-net: learning dense volumetric segmentation from sparse annotation.
\newblock In \emph{Medical Image Computing and Computer-Assisted Intervention--MICCAI 2016: 19th International Conference, Athens, Greece, October 17-21, 2016, Proceedings, Part II 19}, pages 424--432. Springer, 2016.

\bibitem[Dorent et~al.(2023)Dorent, Kujawa, Ivory, Bakas, Rieke, Joutard, Glocker, Cardoso, Modat, Batmanghelich, et~al.]{dorent2023crossmoda}
Reuben Dorent, Aaron Kujawa, Marina Ivory, Spyridon Bakas, Nicola Rieke, Samuel Joutard, Ben Glocker, Jorge Cardoso, Marc Modat, Kayhan Batmanghelich, et~al.
\newblock Crossmoda 2021 challenge: Benchmark of cross-modality domain adaptation techniques for vestibular schwannoma and cochlea segmentation.
\newblock \emph{Medical Image Analysis}, 83:\penalty0 102628, 2023.

\bibitem[Dosovitskiy(2020)]{dosovitskiy2020image}
Alexey Dosovitskiy.
\newblock An image is worth 16x16 words: Transformers for image recognition at scale.
\newblock \emph{arXiv preprint arXiv:2010.11929}, 2020.

\bibitem[Feichtenhofer et~al.(2022)Feichtenhofer, Li, He, et~al.]{feichtenhofer2022masked}
Christoph Feichtenhofer, Yanghao Li, Kaiming He, et~al.
\newblock Masked autoencoders as spatiotemporal learners.
\newblock \emph{Advances in neural information processing systems}, 35:\penalty0 35946--35958, 2022.

\bibitem[Goncharov et~al.(2023)Goncharov, Soboleva, Kurmukov, Pisov, and Belyaev]{voxtovex}
Mikhail Goncharov, Vera Soboleva, Anvar Kurmukov, Maxim Pisov, and Mikhail Belyaev.
\newblock vox2vec: A framework for self-supervised contrastive learning of voxel-level representations in medical images.
\newblock In \emph{Medical Image Computing and Computer Assisted Intervention -- MICCAI 2023}, 2023.

\bibitem[Gr{\o}vik et~al.(2020)Gr{\o}vik, Yi, Iv, Tong, Rubin, and Zaharchuk]{grovik2020deep}
Endre Gr{\o}vik, Darvin Yi, Michael Iv, Elizabeth Tong, Daniel Rubin, and Greg Zaharchuk.
\newblock Deep learning enables automatic detection and segmentation of brain metastases on multisequence mri.
\newblock \emph{Journal of Magnetic Resonance Imaging}, 51\penalty0 (1):\penalty0 175--182, 2020.

\bibitem[Gu et~al.(2024)Gu, Zhang, Li, Wang, and Chen]{gu2024self}
Pengfei Gu, Yejia Zhang, Huimin Li, Chaoli Wang, and Danny~Z Chen.
\newblock Self pre-training with topology-and spatiality-aware masked autoencoders for 3d medical image segmentation.
\newblock \emph{arXiv preprint arXiv:2406.10519}, 2024.

\bibitem[Hatamizadeh et~al.(2021{\natexlab{a}})Hatamizadeh, Nath, Tang, Yang, Roth, and Xu]{hatamizadeh2021swin}
Ali Hatamizadeh, Vishwesh Nath, Yucheng Tang, Dong Yang, Holger~R Roth, and Daguang Xu.
\newblock Swin unetr: Swin transformers for semantic segmentation of brain tumors in mri images.
\newblock In \emph{International MICCAI brainlesion workshop}, pages 272--284. Springer, 2021{\natexlab{a}}.

\bibitem[Hatamizadeh et~al.(2021{\natexlab{b}})Hatamizadeh, Tang, Nath, Yang, Myronenko, Landman, Roth, and Xu]{hatamizadeh2021unetrtransformers3dmedical}
Ali Hatamizadeh, Yucheng Tang, Vishwesh Nath, Dong Yang, Andriy Myronenko, Bennett Landman, Holger Roth, and Daguang Xu.
\newblock Unetr: Transformers for 3d medical image segmentation, 2021{\natexlab{b}}.

\bibitem[He et~al.(2022)He, Chen, Xie, Li, Doll{\'a}r, and Girshick]{he2022masked}
Kaiming He, Xinlei Chen, Saining Xie, Yanghao Li, Piotr Doll{\'a}r, and Ross Girshick.
\newblock Masked autoencoders are scalable vision learners.
\newblock In \emph{Proceedings of the IEEE/CVF conference on computer vision and pattern recognition}, pages 16000--16009, 2022.

\bibitem[He et~al.(2023)He, Yang, Ge, Chen, Coatrieux, Wang, and Li]{he2023geometric}
Yuting He, Guanyu Yang, Rongjun Ge, Yang Chen, Jean-Louis Coatrieux, Boyu Wang, and Shuo Li.
\newblock Geometric visual similarity learning in 3d medical image self-supervised pre-training.
\newblock In \emph{Proceedings of the IEEE/CVF Conference on Computer Vision and Pattern Recognition}, pages 9538--9547, 2023.

\bibitem[Hernandez~Petzsche et~al.(2022)Hernandez~Petzsche, de~la Rosa, Hanning, Wiest, Valenzuela, Reyes, Meyer, Liew, Kofler, Ezhov, et~al.]{hernandez2022isles}
Moritz~R Hernandez~Petzsche, Ezequiel de~la Rosa, Uta Hanning, Roland Wiest, Waldo Valenzuela, Mauricio Reyes, Maria Meyer, Sook-Lei Liew, Florian Kofler, Ivan Ezhov, et~al.
\newblock Isles 2022: A multi-center magnetic resonance imaging stroke lesion segmentation dataset.
\newblock \emph{Scientific data}, 9\penalty0 (1):\penalty0 762, 2022.

\bibitem[Huang et~al.(2023)Huang, Wang, Deng, Ye, Su, Sun, He, Gu, Gu, Zhang, et~al.]{huang2023stu}
Ziyan Huang, Haoyu Wang, Zhongying Deng, Jin Ye, Yanzhou Su, Hui Sun, Junjun He, Yun Gu, Lixu Gu, Shaoting Zhang, et~al.
\newblock Stu-net: Scalable and transferable medical image segmentation models empowered by large-scale supervised pre-training.
\newblock \emph{arXiv preprint arXiv:2304.06716}, 2023.

\bibitem[Isensee et~al.(2021)Isensee, Jaeger, Kohl, Petersen, and Maier-Hein]{isensee2021nnu}
Fabian Isensee, Paul~F Jaeger, Simon~AA Kohl, Jens Petersen, and Klaus~H Maier-Hein.
\newblock nnu-net: a self-configuring method for deep learning-based biomedical image segmentation.
\newblock \emph{Nature methods}, 18\penalty0 (2):\penalty0 203--211, 2021.

\bibitem[Isensee et~al.(2024)Isensee, Wald, Ulrich, Baumgartner, Roy, Maier-Hein, and Jaeger]{isensee2024nnu}
Fabian Isensee, Tassilo Wald, Constantin Ulrich, Michael Baumgartner, Saikat Roy, Klaus Maier-Hein, and Paul~F Jaeger.
\newblock nnu-net revisited: A call for rigorous validation in 3d medical image segmentation.
\newblock \emph{arXiv preprint arXiv:2404.09556}, 2024.

\bibitem[Landman et~al.(2015)Landman, Xu, Igelsias, Styner, and et~al.]{landman20152015}
Bennett Landman, Zhoubing Xu, Juan~Eugenio Igelsias, Martin Styner, and et al.
\newblock 2015 miccai multi-atlas labeling beyond the cranial vault workshop and challenge.
\newblock In \emph{Proc. MICCAI Multi-Atlas Labeling Beyond Cranial Vault—Workshop Challenge}, 2015.

\bibitem[Li et~al.(2023)Li, Fan, Hu, Feichtenhofer, and He]{li2023scaling}
Yanghao Li, Haoqi Fan, Ronghang Hu, Christoph Feichtenhofer, and Kaiming He.
\newblock Scaling language-image pre-training via masking.
\newblock In \emph{Proceedings of the IEEE/CVF Conference on Computer Vision and Pattern Recognition}, pages 23390--23400, 2023.

\bibitem[Liew et~al.(2022)Liew, Lo, Donnelly, Zavaliangos-Petropulu, Jeong, Barisano, Hutton, Simon, Juliano, Suri, et~al.]{liew2022large}
Sook-Lei Liew, Bethany~P Lo, Miranda~R Donnelly, Artemis Zavaliangos-Petropulu, Jessica~N Jeong, Giuseppe Barisano, Alexandre Hutton, Julia~P Simon, Julia~M Juliano, Anisha Suri, et~al.
\newblock A large, curated, open-source stroke neuroimaging dataset to improve lesion segmentation algorithms.
\newblock \emph{Scientific data}, 9\penalty0 (1):\penalty0 320, 2022.

\bibitem[Munk et~al.(2024)Munk, Ambsdorf, Llambias, and Nielsen]{munk2024amaes}
Asbj{\o}rn Munk, Jakob Ambsdorf, Sebastian Llambias, and Mads Nielsen.
\newblock Amaes: Augmented masked autoencoder pretraining on public brain mri data for 3d-native segmentation.
\newblock \emph{arXiv preprint arXiv:2408.00640}, 2024.

\bibitem[Muslim et~al.(2022)Muslim, Mashohor, Gawwam, Mahmud, binti Hanafi, Alnuaimi, Josephine, and Almutairi]{MUSLIM2022108139}
Ali~M. Muslim, Syamsiah Mashohor, Gheyath~Al Gawwam, Rozi Mahmud, Marsyita binti Hanafi, Osama Alnuaimi, Raad Josephine, and Abdullah~Dhaifallah Almutairi.
\newblock Brain mri dataset of multiple sclerosis with consensus manual lesion segmentation and patient meta information.
\newblock \emph{Data in Brief}, 42:\penalty0 108139, 2022.

\bibitem[of~Health et~al.(2015)]{national2015adolescent}
National~Institutes of Health et~al.
\newblock Adolescent brain cognitive development study, 2015.

\bibitem[Oquab et~al.(2023)Oquab, Darcet, Moutakanni, Vo, Szafraniec, Khalidov, Fernandez, Haziza, Massa, El-Nouby, et~al.]{oquab2023dinov2}
Maxime Oquab, Timoth{\'e}e Darcet, Th{\'e}o Moutakanni, Huy Vo, Marc Szafraniec, Vasil Khalidov, Pierre Fernandez, Daniel Haziza, Francisco Massa, Alaaeldin El-Nouby, et~al.
\newblock Dinov2: Learning robust visual features without supervision.
\newblock \emph{arXiv preprint arXiv:2304.07193}, 2023.

\bibitem[Podobnik et~al.(2023)Podobnik, Strojan, Peterlin, Ibragimov, and Vrtovec]{podobnik2023han}
Ga{\v{s}}per Podobnik, Primo{\v{z}} Strojan, Primo{\v{z}} Peterlin, Bulat Ibragimov, and Toma{\v{z}} Vrtovec.
\newblock Han-seg: The head and neck organ-at-risk ct and mr segmentation dataset.
\newblock \emph{Medical physics}, 50\penalty0 (3):\penalty0 1917--1927, 2023.

\bibitem[Ronneberger et~al.(2015)Ronneberger, Fischer, and Brox]{ronneberger2015u}
Olaf Ronneberger, Philipp Fischer, and Thomas Brox.
\newblock U-net: Convolutional networks for biomedical image segmentation.
\newblock In \emph{Medical image computing and computer-assisted intervention--MICCAI 2015: 18th international conference, Munich, Germany, October 5-9, 2015, proceedings, part III 18}, pages 234--241. Springer, 2015.

\bibitem[Roy et~al.(2023)Roy, Koehler, Ulrich, Baumgartner, Petersen, Isensee, J{\"a}ger, and Maier-Hein]{roy}
Saikat Roy, Gregor Koehler, Constantin Ulrich, Michael Baumgartner, Jens Petersen, Fabian Isensee, Paul~F. J{\"a}ger, and Klaus~H. Maier-Hein.
\newblock Mednext: Transformer-driven scaling of convnets for medical image segmentation.
\newblock In \emph{Medical Image Computing and Computer Assisted Intervention -- MICCAI 2023}, 2023.

\bibitem[Simpson et~al.(2019)Simpson, Antonelli, Bakas, Bilello, Farahani, Van~Ginneken, Kopp-Schneider, Landman, Litjens, Menze, et~al.]{simpson2019large}
Amber~L Simpson, Michela Antonelli, Spyridon Bakas, Michel Bilello, Keyvan Farahani, Bram Van~Ginneken, Annette Kopp-Schneider, Bennett~A Landman, Geert Litjens, Bjoern Menze, et~al.
\newblock A large annotated medical image dataset for the development and evaluation of segmentation algorithms.
\newblock \emph{arXiv preprint arXiv:1902.09063}, 2019.

\bibitem[Tang et~al.(2024)Tang, Xu, Yao, Fu, Quan, Zhu, Liu, and Zhou]{tang2024hyspark}
Fenghe Tang, Ronghao Xu, Qingsong Yao, Xueming Fu, Quan Quan, Heqin Zhu, Zaiyi Liu, and S~Kevin Zhou.
\newblock Hyspark: Hybrid sparse masking for large scale medical image pre-training.
\newblock \emph{arXiv preprint arXiv:2408.05815}, 2024.

\bibitem[Tang et~al.(2022)Tang, Yang, Li, Roth, Landman, Xu, Nath, and Hatamizadeh]{tang2022self}
Yucheng Tang, Dong Yang, Wenqi Li, Holger~R Roth, Bennett Landman, Daguang Xu, Vishwesh Nath, and Ali Hatamizadeh.
\newblock Self-supervised pre-training of swin transformers for 3d medical image analysis.
\newblock In \emph{Proceedings of the IEEE/CVF conference on computer vision and pattern recognition}, pages 20730--20740, 2022.

\bibitem[Tian et~al.(2023{\natexlab{a}})Tian, Jiang, Diao, Lin, Wang, and Yuan]{tian2023designing}
Keyu Tian, Yi Jiang, Qishuai Diao, Chen Lin, Liwei Wang, and Zehuan Yuan.
\newblock Designing bert for convolutional networks: Sparse and hierarchical masked modeling.
\newblock \emph{arXiv preprint arXiv:2301.03580}, 2023{\natexlab{a}}.

\bibitem[Tian et~al.(2023{\natexlab{b}})Tian, Jiang, Diao, Lin, Wang, and Yuan]{tian2023designingbertconvolutionalnetworks}
Keyu Tian, Yi Jiang, Qishuai Diao, Chen Lin, Liwei Wang, and Zehuan Yuan.
\newblock Designing bert for convolutional networks: Sparse and hierarchical masked modeling, 2023{\natexlab{b}}.

\bibitem[Tong et~al.(2022)Tong, Song, Wang, and Wang]{tong2022videomae}
Zhan Tong, Yibing Song, Jue Wang, and Limin Wang.
\newblock Videomae: Masked autoencoders are data-efficient learners for self-supervised video pre-training.
\newblock \emph{Advances in neural information processing systems}, 35:\penalty0 10078--10093, 2022.

\bibitem[Ulrich et~al.(2023)Ulrich, Isensee, Wald, Zenk, Baumgartner, and Maier-Hein]{ulrich2023multitalent}
Constantin Ulrich, Fabian Isensee, Tassilo Wald, Maximilian Zenk, Michael Baumgartner, and Klaus~H Maier-Hein.
\newblock Multitalent: A multi-dataset approach to medical image segmentation.
\newblock In \emph{International Conference on Medical Image Computing and Computer-Assisted Intervention}, pages 648--658. Springer, 2023.

\bibitem[Valanarasu et~al.(2023)Valanarasu, Tang, Yang, Xu, Zhao, Li, Patel, Landman, Xu, He, et~al.]{valanarasu2023disruptive}
Jeya Maria~Jose Valanarasu, Yucheng Tang, Dong Yang, Ziyue Xu, Can Zhao, Wenqi Li, Vishal~M Patel, Bennett Landman, Daguang Xu, Yufan He, et~al.
\newblock Disruptive autoencoders: Leveraging low-level features for 3d medical image pre-training.
\newblock \emph{arXiv preprint arXiv:2307.16896}, 2023.

\bibitem[Vaswani(2017)]{vaswani2017attention}
A Vaswani.
\newblock Attention is all you need.
\newblock \emph{Advances in Neural Information Processing Systems}, 2017.

\bibitem[Vincent et~al.(2010)Vincent, Larochelle, Lajoie, Bengio, Manzagol, and Bottou]{vincent2010stacked}
Pascal Vincent, Hugo Larochelle, Isabelle Lajoie, Yoshua Bengio, Pierre-Antoine Manzagol, and L{\'e}on Bottou.
\newblock Stacked denoising autoencoders: Learning useful representations in a deep network with a local denoising criterion.
\newblock \emph{Journal of machine learning research}, 11\penalty0 (12), 2010.

\bibitem[Wahid et~al.(2024)Wahid, Dede, Naser, and Fuller]{hntsmrg2024wahid}
Kareem Wahid, Cem Dede, Mohamed Naser, and Clifton Fuller.
\newblock Training dataset for hntsmrg 2024 challenge, 2024.

\bibitem[Wang et~al.(2023)Wang, Wu, Luo, Liu, Li, and Zhang]{wang2023mis}
Guotai Wang, Jianghao Wu, Xiangde Luo, Xinglong Liu, Kang Li, and Shaoting Zhang.
\newblock Mis-fm: 3d medical image segmentation using foundation models pretrained on a large-scale unannotated dataset.
\newblock \emph{arXiv preprint arXiv:2306.16925}, 2023.

\bibitem[Wasserthal et~al.(2023)Wasserthal, Breit, Meyer, Pradella, Hinck, Sauter, Heye, Boll, Cyriac, Yang, et~al.]{wasserthal2023totalsegmentator}
Jakob Wasserthal, Hanns-Christian Breit, Manfred~T Meyer, Maurice Pradella, Daniel Hinck, Alexander~W Sauter, Tobias Heye, Daniel~T Boll, Joshy Cyriac, Shan Yang, et~al.
\newblock Totalsegmentator: robust segmentation of 104 anatomic structures in ct images.
\newblock \emph{Radiology: Artificial Intelligence}, 5\penalty0 (5), 2023.

\bibitem[Woo et~al.(2023)Woo, Debnath, Hu, Chen, Liu, Kweon, and Xie]{woo2023convnext}
Sanghyun Woo, Shoubhik Debnath, Ronghang Hu, Xinlei Chen, Zhuang Liu, In~So Kweon, and Saining Xie.
\newblock Convnext v2: Co-designing and scaling convnets with masked autoencoders.
\newblock In \emph{Proceedings of the IEEE/CVF Conference on Computer Vision and Pattern Recognition}, pages 16133--16142, 2023.

\bibitem[Wu et~al.(2024)Wu, Zhuang, and Chen]{wu2024voco}
Linshan Wu, Jiaxin Zhuang, and Hao Chen.
\newblock Voco: A simple-yet-effective volume contrastive learning framework for 3d medical image analysis.
\newblock In \emph{Proceedings of the IEEE/CVF Conference on Computer Vision and Pattern Recognition}, pages 22873--22882, 2024.

\bibitem[Zhang et~al.(2024)Zhang, Wu, Angelini, Li, Guo, Rasmussen, O'Connor, Wadhwa, Jackowski, Li, Posner, Laine, and Wang]{zhang2024mapsegunifiedunsuperviseddomain}
Xuzhe Zhang, Yuhao Wu, Elsa Angelini, Ang Li, Jia Guo, Jerod~M. Rasmussen, Thomas~G. O'Connor, Pathik~D. Wadhwa, Andrea~Parolin Jackowski, Hai Li, Jonathan Posner, Andrew~F. Laine, and Yun Wang.
\newblock Mapseg: Unified unsupervised domain adaptation for heterogeneous medical image segmentation based on 3d masked autoencoding and pseudo-labeling, 2024.

\bibitem[Zhou et~al.(2021)Zhou, Sodha, Pang, Gotway, and Liang]{zhou2021models}
Zongwei Zhou, Vatsal Sodha, Jiaxuan Pang, Michael~B Gotway, and Jianming Liang.
\newblock Models genesis.
\newblock \emph{Medical image analysis}, 67:\penalty0 101840, 2021.

\bibitem[Zhuang et~al.(2023)Zhuang, Luo, and Chen]{zhuang2023advancing}
Jia-Xin Zhuang, Luyang Luo, and Hao Chen.
\newblock Advancing volumetric medical image segmentation via global-local masked autoencoder.
\newblock \emph{arXiv preprint arXiv:2306.08913}, 2023.

\bibitem[Zhuang et~al.(2019)Zhuang, Li, Hu, Ma, Yang, and Zheng]{zhuang2019self}
Xinrui Zhuang, Yuexiang Li, Yifan Hu, Kai Ma, Yujiu Yang, and Yefeng Zheng.
\newblock Self-supervised feature learning for 3d medical images by playing a rubik’s cube.
\newblock In \emph{Medical Image Computing and Computer Assisted Intervention--MICCAI 2019: 22nd International Conference, Shenzhen, China, October 13--17, 2019, Proceedings, Part IV 22}, pages 420--428. Springer, 2019.

\end{thebibliography}
}
\newpage
\clearpage
\appendix
\section{Method configuration}
Across all baseline methods, we utilize a common set of hyperparameters.
For all baseline methods we utilize the same pre-training dataset with the same image preprocessing. Moreover we use the same amount of pre-training steps (250k) as for our S3D-B method and the same fine-tuning scheme, as highlighted in \cref{tab:finetune_results_table}. Aside from this, we employ the SGD optimizer with LR 1e-2 with a PolyLR schedule, momentum 0.99 and weight decay 3e-5 across all pre-training experiments, as they showed to be highly robust and reliable in the supervised medical image segmentation setting using CNNs \citep{isensee2021nnu}. 
Moreover, we denote that all these methods have their backbones replaced with a ResEnc U-Net to minimize confounding effects of different architectures.

\label{apx:detailed_method_config}
\subsection{Models Genesis}
\label{apx:models_genesis}
Models genesis \citep{zhou2021models} pre-text task is centered around restoring original patches from transformed versions. The transformed version is achieved by applying four different transformations in various combinations, with the following transformations:
a composition of four separate pre-training schemes:
\begin{enumerate*}[label=(\roman*)]
\item \textbf{Non-linear intensity transformation:} Alters the intensity distribution while preserving the anatomy, focusing on learning the appearance of organs.
\item \textbf{Out-painting:} Removes part of the image and requires the model to extrapolate from the remaining image, forcing it to learn the global structure of the organs.
\item \textbf{In-painting:} Masks a part of the image, and the model learns to restore the missing parts, focusing on local continuity and context.
\end{enumerate*}
After having transformed the original image through these augmentations, the model is trained to recover the original image through a convolutional encoder-decoder architecture. This approach consolidates different tasks (appearance, texture, and context learning) into one unified image restoration task, making the model more robust and generalizable.

\paragraph{Model specific Hyperparameters:}
The entire set of hyperparameters of Models Genesis are contained within the data-augmentation. This allows us to transfer this transformation pipeline, as provided in the official \href{https://github.com/MrGiovanni/ModelsGenesis}{repository} without any changes to the hyperparameters.

\subsection{VolumeFusion}
\label{apx:volume_fusion}
Volume Fusion \citep{wang2023mis} is a pseudo-segmentation task using two sub-volumes from different 3D scans, which are fused together based on random voxel-level fusion coefficients.
The fused image is treated as input, and the model predicts the fusion category of each voxel, mimicking a segmentation task.
Pretraining is optimized using a combination of Dice loss and cross-entropy loss.

\paragraph{Method specific parameters:} Volume Fusion has unique parameters defining the size ranges of the rectangles used for fusing together images. In our experiments we utilize a rectangle size range between [8, 100] sampled uniformly for each axis. This represent the 62.5\% of our input patch size, and identical percentage as in the original paper. 
Moreover the amount of rectangles sampled is an important parameter. Like in the original paper we sample $M \sim \mathcal{U}(10, 40)$ different rectangles, iteratively. Lastly, the number of categories was chosen to be 5, as in the original paper (this represent \textit{K} = 4). 

\subsection{VoCo}
\label{apx:voco}
The 'Volume Contrastive Learning Framework' (VoCo) \citep{wu2024voco} is designed to enhance self-supervised learning for 3D medical image analysis by leveraging the consistent contextual positions of anatomical structures. 
The method involves generating base crops from different regions of 3D images and using these as class assignments.
The framework then contrasts random sub-volume crops against these base crops, predicting their contextual positions using a contrastive learning approach.
The authors utilize a Swin-UNETR model architecture, employing the AdamW optimizer with a cosine learning rate schedule for 100,000 pre-training steps. The specific hyperparameters include cropping non-overlapping volumes with a size of 64x64x64, and generating 4x4 base crops during the position prediction task. This represents an input patch size of 384×384×96 which is rescaled and resized to fit exactly 4x4 64x64x64 crops. 

\noindent Since our chosen patch size 160x160x160 is incompatible with the 64 cube length, we adjusted our patch size for VoCo to 192x192x64. This accommodates a 3x3 grid of 64x64x64. Unfortunately the 4x4 grid led to exceeding the memory limit hence a reduction was necessary. Moreover we increased the target crop size from 4 originally to 5 and increased the batch size from 6 (default in our other experiments) to 12, to fully utilize the 40GB VRAM of an A100 node.


\section{Longer Training schedule}
MAEs are known to benefit from increasing the length of the training schedule, as shown in \citet{he2022masked}. We evaluate if this effect transfers to 3D medical pre-training by increasing the training batch size by x8 to $48$, the learning rate to $3e-2$, and the iteration steps by x5 to a total of $1.25M$ steps. We refer to this model as \textbf{S3D-Long} to denote the longer training schedule with more data seen. We denote that the architecture remains identical to the previous architecture, to isolate the effect of the length of the steps as well as the amount of samples seen. 
Results are presented in \cref{tab:placeholderscaling}. It can be observed that this x32 actually leads to a decrease in overall model performance on our test datasets, showing a 0.6\% lower average DSC as well as a 0.6\% lower Average NSD. 

\noindent The observed performance degradation of the S3D-Long model, despite the longer training schedule and increased data exposure, suggests several possible factors at play. While MAEs have shown benefits from extended training schedules in general computer vision tasks, the same assumptions may not directly transfer to 3D medical image pre-training due to the unique nature of this domain. The findings highlight the importance of tailoring training strategies to the domain. 
\begin{table}
            \centering
            \caption{\textbf{Publicly available checkpoint trained on the ABCD dataset:} To provide a public available checkpoint, we retrained our proposed model on the ABCD dataset, indicated by a *. It performs slightly worse than the network pre-trained on the private dataset. }
            \resizebox{\linewidth}{!}{%
\begin{tabular}{l|cccc}
\toprule
SSL Method &            No (Dyn.) &           No (Fix.) &                  \textbf{S3D*} &                \textbf{S3D} \\
\midrule
\textbf{Dataset}              &    \multicolumn{4}{c}{\textbf{Dice Similarity Coefficient (DSC)}}\\
\midrule
\rowcolor{lightgray!30}MS FLAIR (D1)        &              57.81 &  \underline{59.82} &   59.75    &     \textbf{60.35} \\
Brain Mets (D2)      &              63.66 &              56.53 &  \underline{64.20}    &  \textbf{65.24} \\
\rowcolor{lightgray!30}Hippocampus (D3)     &              89.18 &              89.24 & \underline{89.45}&     \textbf{89.60} \\
Atlas22 (D4)         &              63.28 &              65.52 & \underline{66.61} &     \textbf{66.95} \\
\rowcolor{lightgray!30}CrossModa (D5)       &     \textbf{85.64} &              83.44 & 83.61 &              \underline{84.08} \\
Cosmos22 (D6)        &               60.28 &              78.17 & \textbf{80.01}  &  \underline{80.00} \\
\rowcolor{lightgray!30}ISLES22 (D7)         &              77.94 &  \underline{79.44} & 78.94 &     \textbf{79.70} \\
Hanseg (D8)          &              59.00 &              \underline{61.85} & 61.27  &  \textbf{62.11} \\
\rowcolor{lightgray!30}HNTS-MRG24 (D9)      &              66.73 &              65.90 & \underline{67.03}&     \textbf{68.62} \\
BRATS24 Africa (D10) &     \textbf{93.07} &  \underline{92.51} & 92.49 &              92.19 \\\midrule
\textbf{Avg. DSC} & 71.66 & 73.24 & \underline{74.34} & \textbf{74.88} \\
\textbf{Avg. Rank} & 3.2 & 2.9 & \underline{2.4} & \textbf{1.5} \\
\midrule
\end{tabular}}
            \label{subtab:abcdresults} 
\end{table}

\section{Additional results}
\label{apx:finetune_multi_channel}
Aside from the quantitative data on the development and test dataset, we provide the quantiative data of the ablation experiments here. 
The following additional results are provided:
\begin{enumerate*}
    \item Results when fine-tuning in a low-data regime are presented in \cref{tab:low_data}.
    \item Experiment on how to best transfer weights when transferring to a dataset with more than 1 input channel is provided in  \cref{tab:stemtable}
    \item Results on how the pre-training effects generalization is provided in \cref{tab:generalization}.
    \item Experiment results of  investigating if one can reduce the fine-tuning steps are presented in \cref{tab:finetune_results_table}
\end{enumerate*}   

\subsection{Public weights trained on the ABCD dataset} Due to patient privacy concerns and data ownership regulations, we are unable to share the original pre-trained weights. As an alternative, we retrained our best-performing model on the National Institute of Health's Adolescent Brain Cognitive Development (ABCD) dataset. This dataset comprises about 41k MRI scans with a 50-to-50 ratio of T1-weighted to T2-weighted scans. The results, shown in \cref{subtab:abcdresults}, indicate that the original model slightly outperforms the version pre-trained on the ABCD dataset. This discrepancy is likely attributable to the greater diversity and variation in the images within our private dataset, which enables a more robust feature representation.

\subsection{Comparison to previous work on CT data}
Although this study focuses on brain MRI images, we also evaluate our method on a CT downstream task. \Cref{tab:btcvresults} presents the results of our approach on the BTCV multi-organ segmentation task \cite{landman20152015}. For comparison, we incorporate a diverse set of results reported in \citep{tang2024hyspark}, which were trained using an unspecified 80/20 data split. Additionally, we fine-tune the publicly available HySpark checkpoint \citep{tang2024hyspark} using the same five-fold cross-validation split as for our method. Remarkably, M3D outperforms all other methods, despite being pretrained exclusively on brain MRI data. Notably, none of the related approaches surpass our backbone trained on scratch. This highlights the critical role of leveraging state-of-the-art networks and advanced training frameworks, such as nnU-Net. \cite{isensee2021nnu, isensee2024nnu}.

\begin{table}[]
    \centering
    \caption{\textbf{Forty images with SSL are almost as good as all data from-scratch!} The pre-trained S3D model almost reaches the performance of the model trained from-scratch with only 40 training cases, with the exception of D4. Overall train/val/test dataset size was 38/10/12 for D1, 67/17/21 for D2, 166/42/52 for D3, 419/105/131 for D4, 134/34/42 for D5. Results in the table are reported on the validation set. \textit{full: Uses all train samples of the dataset. * D1 has only 38 training cases for the train split.}}
    \label{tab:low_data}
    \resizebox{\linewidth}{!}{%
\begin{tabular}{llrrrrrr}
\toprule
SSL Method & N Train &    D1 & D2 &  D3 &  D4&  D5 &  Avg. D1-D5 \\
\midrule
\multirow{5}{*}{Scratch} & 10\cellcolor{lightgray!30} &    \cellcolor{lightgray!30}      40.78 &  \cellcolor{lightgray!30}          43.52 &        \cellcolor{lightgray!30}     84.94 &    \cellcolor{lightgray!30}     44.11 &     \cellcolor{lightgray!30}      76.66 &   \cellcolor{lightgray!30}    58.00 \\
             & 20 &          44.46 &            59.46 &             86.75 &         46.33 &           78.67 &       63.13 \\
           &   30 \cellcolor{lightgray!30}&      \cellcolor{lightgray!30}    45.42 &      \cellcolor{lightgray!30}      64.20 &          \cellcolor{lightgray!30}   87.14 &      \cellcolor{lightgray!30}   48.22 &      \cellcolor{lightgray!30}     78.47 &   \cellcolor{lightgray!30}    64.69 \\
             & 40 &          49.37* &            60.13 &             87.59 &         50.43 &           78.37 &       65.18 \\
         &  full\cellcolor{lightgray!30} &         \cellcolor{lightgray!30} 49.37 &    \cellcolor{lightgray!30}        69.13 &       \cellcolor{lightgray!30}      88.78 &     \cellcolor{lightgray!30}    60.74 &         \cellcolor{lightgray!30}  81.33 &     \cellcolor{lightgray!30}  69.87 \\\midrule
\multirow{5}{*}{S3D (ours)} & 10 \cellcolor{lightgray!30} &     \cellcolor{lightgray!30}     43.48 &      \cellcolor{lightgray!30}      48.44 &       \cellcolor{lightgray!30}      84.12 &     \cellcolor{lightgray!30}    41.51 &    \cellcolor{lightgray!30}       77.70 &   \cellcolor{lightgray!30}    59.05 \\
             & 20 &          46.58 &            65.30 &             86.61 &         45.50 &           79.52 &       64.70 \\
           & 30 \cellcolor{lightgray!30}&       \cellcolor{lightgray!30}   48.12 &      \cellcolor{lightgray!30}      68.41 &    \cellcolor{lightgray!30}         86.77 &    \cellcolor{lightgray!30}     51.62 &   \cellcolor{lightgray!30}        78.88 &   \cellcolor{lightgray!30}    66.76 \\
             &  40 &          51.49* &            72.91 &             87.46 &         53.05 &           80.82 &       69.15 \\
      & full \cellcolor{lightgray!30}&     \cellcolor{lightgray!30}     51.49 &    \cellcolor{lightgray!30}        74.01 &       \cellcolor{lightgray!30}      88.83 &     \cellcolor{lightgray!30}    62.39 &     \cellcolor{lightgray!30}      81.54 &   \cellcolor{lightgray!30}    71.65 \\
\bottomrule
\end{tabular}}

\end{table}

\begin{table}[t]
    \centering 
    \caption{\textbf{Pre-training length ablation:} Longer pre-training does not lead to improved performance. Interestingly, when exceeding 250k steps.}
    \label{tab:pretrainind_length}
    \resizebox{\linewidth}{!}{%
\begin{tabular}{lrrrrrrr}
\toprule
PT Iterations &    D1 &  D2 &  D3 &  D4 &  D5 &  Avg. D1-D5 &  Train Time [h] \\
\midrule
62.5k       &          49.49 &            70.79 &             88.82 &         62.95 &           81.27 &       70.67 &              28 \\
125k       &          50.56 &            70.48 &             88.86 &         62.51 &           81.69 &       70.82 &              56 \\
\rowcolor{lightgray!60} 250k      &          51.02 &            74.07 &             88.91 &         62.81 &           81.50 &       71.66 &             112 \\
500k      &          50.93 &            72.71 &             88.88 &         62.17 &           81.86 &       71.31 &             224 \\
1M      &          50.45 &            71.55 &             88.92 &         62.78 &           81.82 &       71.10 &             448 \\
\bottomrule
\end{tabular}}

\end{table}

\begin{table}[]
    \centering
    \caption{Replicating the pre-trained stem weights and freezing them during the decoder warm-up phase yields the most stable and equally best results. }
    \label{tab:stemtable}
    \resizebox{\linewidth}{!}{%
\begin{tabular}{llccccc|cc}
\toprule
Initialization & Decoder Warm-Up & Fold 0 & Fold 1 & Fold 2 & Fold 3 & Fold 4 & Average & STD  \\
\midrule
Replication      & Frozen              & 72.84  & 64.42  & 66.11  & 62.86  & 62.85  & \underline{65.82}   & 4.15 \\
Replication      & Unfrozen            & 72.68  & 63.07  & 65.60  & 66.02  & 61.08  & 65.69   & 4.39\\
Random         & Frozen              & 74.38  & 60.89  & 65.10  & 67.55  & 61.31  & \textbf{65.85 }  & 5.51 \\
Random         & Unfrozen            & 72.20  & 63.16  & 62.25  & 66.71  & 61.47  & 65.16   & 4.42 \\
\bottomrule
\end{tabular}}

\end{table}

\begin{table}
    \centering
    \caption{\textbf{Pre-training can improve generalization:} We investigate generalization to a new modality time-of-flight (ToF) MRI (top), and the generalization of a resulting method when translating it to a different clinic (bottom).
    }
    \label{tab:generalization}
    \resizebox{\linewidth}{!}{
\begin{tabular}{ll|rrrrrr}
\toprule
Experiment & Setting &  No Dyn. &  No Fixed &  VoCo &    VF &    MG &  S3D-B  \\
\midrule
Modality shift & TOF Angio. Aneurysms(D12) &    \textbf{42.61} &     22.76 & 22.32 & 31.21 & \underline{34.60} &         28.72\\
\midrule
In Distribution & Brain Mets (D2) &    \textbf{72.81} &     67.93 & 64.34 & \underline{71.69} & 69.05 &         71.56 \\
Patient shift & Brain Mets (D13) &    64.08 &     61.61 & 56.78 & 63.95 & \underline{64.22} &         \textbf{64.54} \\
\bottomrule
\end{tabular}}

\end{table}

\begin{table}
    \centering
    \caption{\textbf{Fine-tuning length:} When initializing from our pre-trained checkpoint, it is possible to achieve a large fraction of the final performance after less than 15\% of the normal training time. Despite this a full training schedule reaches better performance. These experiments were conducted using S3D long on the validation splits.}
    \resizebox{\linewidth}{!}{%
\begin{tabular}{lrrrrrr}
\toprule
FT Iterations &   D1 &  D2 &  D3 &  D4 &  D5 &  Avg. D1-D5 \\
\midrule
\rowcolor{lightgray!30}25k         &          50.85 &            73.99 &             88.51 &         55.49 &           46.00 &       62.97 \\
37.5k        &          51.69 &            74.03 &             88.85 &         60.22 &           81.68 &       71.29 \\
\rowcolor{lightgray!30} 50k       &          51.13 &            73.53 &             88.93 &         60.14 &           81.92 &       71.13 \\
75k       &          51.41 &            72.80 &             89.08 &         63.14 &           81.83 &       71.65 \\
\rowcolor{lightgray!30}150k       &          50.95 &            71.28 &             88.96 &         62.51 &           81.92 &       71.13 \\
275k      &          53.10 &            71.24 &             89.14 &         63.55 &           82.53 &       71.91 \\
\bottomrule
\end{tabular}}

    \label{tab:finetune_results_table}
    
\end{table}
    \begin{table*}[]
    \centering
    \caption{\textbf{S3D outperforms all related work on the BTCV dataset \citep{landman20152015}.} Despite being pretrained exclusively on brain MRI data, our network outperforms all related methods. Values above the line are sourced from \citet{tang2024hyspark}, which used an unknown 80/20 split. To  fine-tuned their published pre-trained weights using a 5-fold cross-validation. Below the line, all models, including ours, were trained on the same 5-fold cross-validation. Notably, even though all related work leveraged CT data for pretraining, none surpassed the performance of our backbone model trained from scratch, emphasizing the importance of Pitfall 2. }
    \label{tab:btcvresults}
    \resizebox{\linewidth}{!}{
        \centering
        \begin{tabular}{llrrrrrrrrrrrr}
        \toprule
        \textbf{Pre-training Method} & \textbf{Network} & \textbf{Spl} & \textbf{Kid} & \textbf{Gall} & \textbf{Eso} & \textbf{Liv} & \textbf{Sto} & \textbf{Aor} & \textbf{IVC} & \textbf{Veins} & \textbf{Pan} & \textbf{AG} & \textbf{Avg} \\
        \hline
        vox2vec~\cite{voxtovex} & 3D UNet(FPN)~\cite{ronneberger2015u} & 91.40 & \textbf{90.70} & 59.50 & 72.70 & \textbf{96.30} & 83.20 & 91.30 & 83.90 & 69.20 & 73.90 & 65.20 & 79.50 \\
        SUP~\cite{hatamizadeh2021swin} & Swin UNETR~\cite{hatamizadeh2021swin} & 84.20 & 86.70 & 58.40 & 70.40 & 94.40 & 76.00 & 87.70 & 82.10 & 67.00 & 69.80 & 61.00 & 75.80 \\
        MAE~\cite{he2022masked} & UNETR~\cite{hatamizadeh2021unetrtransformers3dmedical} & 90.71 & 87.63 & 62.50 & 72.60 & \underline{96.09} & \textbf{94.73} & 86.11 & \textbf{90.36} & 71.00 & 75.47 & 63.77 & 79.07 \\
        SimMIM~\cite{tian2023designingbertconvolutionalnetworks} & Swin UNETR~\cite{hatamizadeh2021swin} & 88.33 & 86.82 & 62.43 & 74.36 & 92.35 & 90.70 & 83.03 & \underline{87.43} & 68.04 & 68.43 & 58.65 & 76.44 \\
        SparK~\cite{tian2023designingbertconvolutionalnetworks} & MedNeXt~\cite{roy} & 90.92 & 87.66 & 62.43 & 74.36 & 95.03 & 84.85 & 86.04 & 80.63 & 68.83 & 76.57 & 61.43 & 79.21 \\
        HySparK~\cite{tang2024hyspark} & MedNeXt+ViT ~\cite{tang2024hyspark}& 90.67 & 88.32 & 68.18 & 74.20 & 95.03 & 87.46 & 90.17 & 84.50 & 70.04 & 78.36 & 66.75 & 80.67 \\
        \midrule
        No & MedNeXt+ViT ~\cite{tang2024hyspark}& 90.35 & 87.46 & 63.18 & 74.49 & 95.09 & 86.00 & 89.29 & 83.22 & 71.85 & 79.48 & 62.59 & 80.27 \\
        HySparK~\cite{tang2024hyspark}& MedNeXt+ViT~\cite{tang2024hyspark} & 90.94 & 86.99 & 63.43 & 74.39 & 95.12 & 87.15 & 88.92 & 83.48 & 72.77 & 79.66 & 64.84 & 80.67 \\
        No (Dyn.) & nnU-Net \cite{isensee2021nnu}  & 90.44 & 88.52 & \underline{68.86} & 78.14 & 95.53 & 88.06 & 91.59 & 86.47 & 76.27 & 81.78 & 71.06 & 83.34 \\
        No (Fix.) & ResEncL (fixed) \cite{isensee2024nnu}& \underline{91.97} & 89.58 & 68.76 & \textbf{79.18} & 95.96 & 91.97 & \underline{92.80} & 87.16 & \textbf{77.29} & \underline{84.01} & \underline{72.21} & \underline{84.63} \\
        S3D& ResEncL (fixed) \cite{isensee2024nnu} & \textbf{92.00} & \underline{90.40} & \textbf{70.77} & \underline{78.71} & 96.01 & \underline{92.51} & \textbf{92.83 }& 87.04 & \underline{77.28} & \textbf{84.79} & \textbf{72.58} & \textbf{84.99} \\
        \bottomrule
    \end{tabular}}

\end{table*}

\begin{table}[]
    \centering
    \caption{\textbf{Longer training schedule degrades performance.} When training with a larger batch size, higher learning rate, and more train steps we observe a degradation in performance for DSC and NSD. Ranks are calculated only between the four methods presented in the table.}
    \label{tab:placeholderscaling}
    \resizebox{\linewidth}{!}{
\begin{tabular}{l|cc|cc}
\toprule
 & \multicolumn{4}{c}{Dice Similarity Coefficient} \\

Dataset &No Dyn. & No Fixed & S3D & S3D-Long \\
\midrule
\rowcolor{lightgray!30}MS FLAIR (D1) & 57.81 & 59.82 & \textbf{60.35} & \underline{59.85} \\
Brain Mets (D2) & 63.66 & 56.53 & \textbf{65.24} & \underline{64.81} \\
\rowcolor{lightgray!30}Hippocampus (D3) & 89.18 & 89.24 & \textbf{89.60} & \underline{89.34} \\
Atlas22 (D4) & 63.28 & \underline{65.52} & \textbf{66.95} & 64.58 \\
\rowcolor{lightgray!30}CrossModa (D5) & \textbf{85.64} & 83.44 & \underline{84.08} & 84.02 \\
Cosmos22 (D6) & 60.28 & 78.17 & \underline{80.00} & \textbf{80.01} \\
\rowcolor{lightgray!30}ISLES22 (D7) & 77.94 & 79.44 & \underline{79.70} & \textbf{79.89} \\
Hanseg (D8) & 59.00 & 61.85 & \textbf{62.11} & \underline{61.93} \\
\rowcolor{lightgray!30}HNTS-MRG24 (D9) & 66.73 & 65.90 & \textbf{68.62} & \underline{67.94} \\
BRATS24 Africa (D10) & \textbf{93.07} & 92.51 & 92.19 & \underline{92.90} \\
\rowcolor{lightgray!30}T2 Aneurysms (D11) & \underline{46.76} & 41.97 & \textbf{47.26} & 44.15 \\
\midrule
Avg. DSC & 69.40 & 70.40 & \textbf{72.37} & \underline{71.77} \\
Avg. Rank & 3.09 & 3.27 & \textbf{1.55} & \underline{2.09} \\
\bottomrule
 & \multicolumn{4}{c}{Normalized Surface Distance} \\
\midrule
\rowcolor{lightgray!30}MS FLAIR (D1) & 78.77 & \underline{80.16} & 80.03 & \textbf{80.40} \\
Brain Mets (D2) & 80.72 & 76.72 & \textbf{82.53} & \underline{82.32} \\
\rowcolor{lightgray!30}Hippocampus (D3) & \textbf{99.46} & 99.42 & \underline{99.46} & 99.44 \\
Atlas22 (D4) & 70.52 & \underline{73.77} & \textbf{75.35} & 73.45 \\
\rowcolor{lightgray!30}CrossModa (D5) & \textbf{99.85} & 99.76 & \underline{99.81} & 99.80 \\
Cosmos22 (D6) & 72.60 & 96.47 & \textbf{97.45} & \underline{96.75} \\
\rowcolor{lightgray!30}ISLES22 (D7) & 88.55 & 90.45 & \underline{90.59} & \textbf{90.72} \\
Hanseg (D8) & 82.20 & \underline{85.94} & 85.80 & \textbf{86.20} \\
\rowcolor{lightgray!30}HNTS-MRG24 (D9) & 71.83 & 71.26 & \textbf{74.07} & \underline{73.17} \\
BRATS24 Africa (D10) & \underline{95.66} & 95.36 & 95.06 & \textbf{95.72} \\
\rowcolor{lightgray!30}T2 Aneurysms (D11) & \textbf{62.24} & 55.56 & \underline{61.18} & 57.07 \\
\midrule
Avg. NSD & 82.04 & 84.08 & \textbf{85.58} & \underline{85.00} \\
Avg. NSD Rank & \underline{2.82} & 3.18 & \textbf{2.00} & \textbf{2.00} \\
\bottomrule
\end{tabular}}

\end{table}

\begin{figure*}
    \centering
    \includegraphics[width=.85\linewidth]{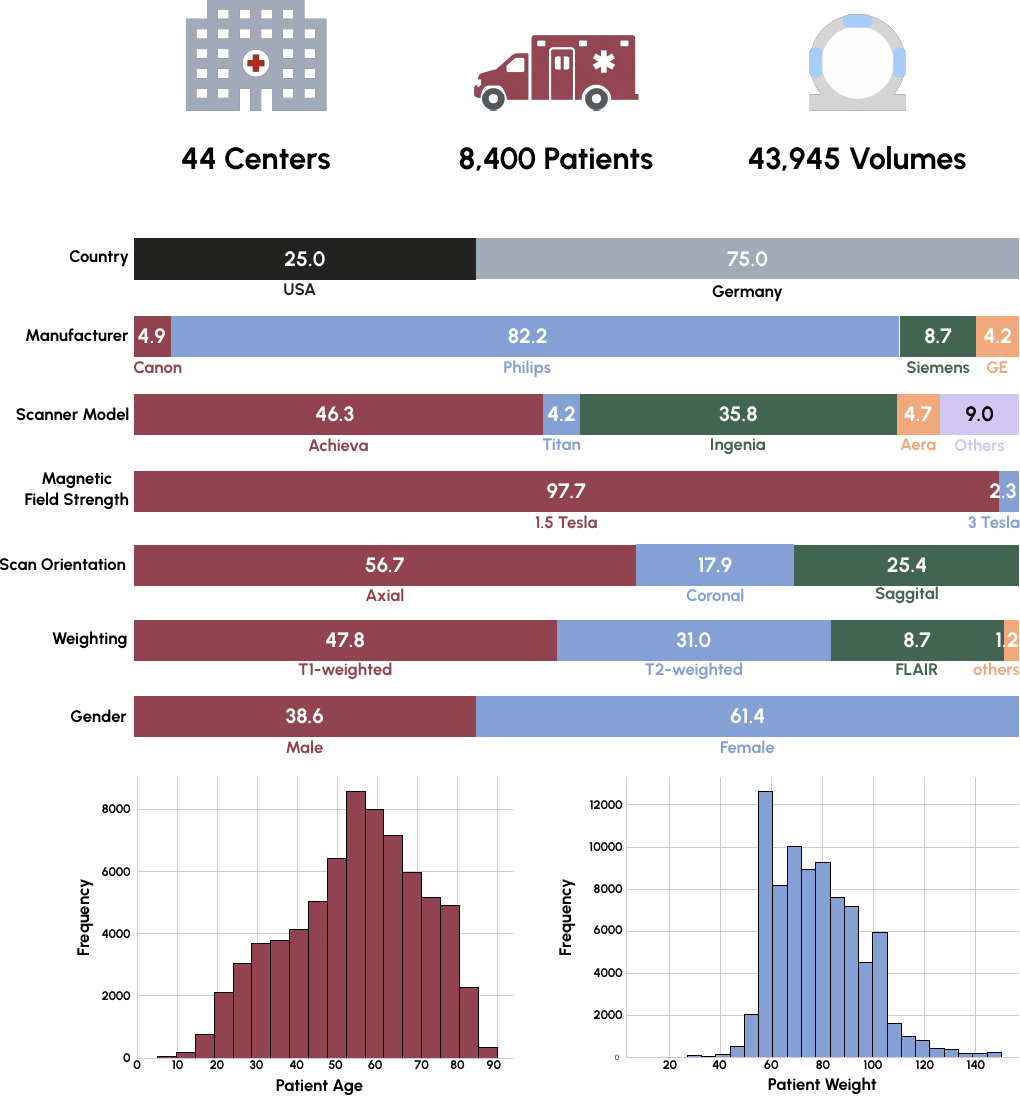}
    \caption{Distribution of our pre-training dataset. The dataset stems from 44 centers and includes 8400 Patients with a 60 to 40 female-to-male ratio. Most patients were imaged with a 1.5 Tesla Philips Achieva or Ingenia scanner. The most prevalent modalities are T1 and T2-weighted images with some additional FLAIR images present. While other modalities were in the dataset, these were not used as prevalence was deemed too low.}
    \label{fig:datasetfigure}
\end{figure*}

\section{Distinction to AMAEs}
Concurrently with this work, \citep{munk2024amaes} introduced the AMAEs framework. Like our approach, it utilizes a dataset of approximately 40k 3D images and employs a state-of-the-art CNN architecture. However, in contrast to our more comprehensive evaluation, their assessment is based on three in-distribution and one out-of-distribution downstream dataset.\\

\noindent While evaluation on three in and one out-of-distribution datasets can suffice to draw some insights, their evaluation setup has additional limitations. First, they constrain themselves to a low-data regime and do not assess whether their performance surpasses a default nnU-Net baseline that is trained for 1000 epochs. Second, their fine-tuning strategy is fixed to a single-channel input to align with the pre-trained stem weights.
While this would be okay for their method, they apply the same limitation to their baselines, which typically utilize all available modalities, which can lead to stronger performance. For instance, in their evaluation on the BraTS dataset, only one of the four available modalities was used, potentially limiting the effectiveness of the baseline models. These choices may impact result reliability and align with Pitfall 3—unreliable evaluation practices. To address such concerns, our work adopts a very comprehensive evaluation strategy to allow drawing reliable conclusions on the state-of-the-art in self-supervised learning for 3D medical image segmentation for the first time. 

\section{Acknowledgement of ABCD}
Data used in the preparation of this article were obtained from the Adolescent Brain Cognitive Development\textsuperscript{SM} (ABCD) Study (\href{https://abcdstudy.org}{https://abcdstudy.org}), held in the NIMH Data Archive (NDA). This is a multisite, longitudinal study designed to recruit more than 10,000 children age 9-10 and follow them over 10 years into early adulthood.
The ABCD Study® is supported by the National Institutes of Health and additional federal partners under award numbers U01DA041048, U01DA050989, U01DA051016, U01DA041022, U01DA051018, U01DA051037, U01DA050987, U01DA041174, U01DA041106, U01DA041117, U01DA041028, U01DA041134, U01DA050988, U01DA051039, U01DA041156, U01DA041025, U01DA041120, U01DA051038, U01DA041148, U01DA041093, U01DA041089, U24DA041123, U24DA041147. A full list of supporters is available at \href{https://abcdstudy.org/federal-partners.html}{https://abcdstudy.org/federal-partners.html}.
A listing of participating sites and a complete listing of the study investigators can be found at \href{https://abcdstudy.org/consortium_members/}{https://abcdstudy.org/consortium\_members/}.
ABCD consortium investigators designed and implemented the study and/or provided data but did not necessarily participate in the analysis or writing of this report. This manuscript reflects the views of the authors and may not reflect the opinions or views of the NIH or ABCD consortium investigators.

\end{document}